\documentclass{article}

\usepackage{microtype}
\usepackage{graphicx}
\usepackage{subcaption}
\usepackage{booktabs}

\usepackage{hyperref}

\usepackage[accepted]{icml2023}

\usepackage{amsmath}
\usepackage{amssymb}
\usepackage{mathtools}
\usepackage{amsthm}
\usepackage{adjustbox}
\usepackage{threeparttable,siunitx}
\usepackage{etoolbox}
\usepackage{multirow}
\usepackage{multicol}
\usepackage{enumerate}

\theoremstyle{plain}
\newtheorem{theorem}{Theorem}
\newtheorem{lemma}{Lemma}
\newtheorem{corollary}{Corollary}
\newtheorem{definition}{Definition}

\def\gA{{\mathcal{A}}}
\def\gD{{\mathcal{D}}}
\def\gS{{\mathcal{S}}}
\def\gM{{\mathcal{M}}}
\def\gN{{\mathcal{N}}}

\newcommand{\bp}{\pi_{\beta}}

\newcommand{\sa}{s,a}

\newcommand{\pare}[1][\sa]{\left( #1 \right)}

\newcommand{\lP}{\widehat{P}}

\newcommand{\lM}{\widehat{M}}
\newcommand{\pM}{\widetilde{M}}

\newcommand{\lR}{\hat{r}}
\newcommand{\pR}{\tilde{r}}

\newcommand{\cnt}[1][\sa]{n\pare[#1]}
\newcommand{\hcnt}[1][\sa]{\hat{n}\pare}

\DeclareMathOperator*{\argmax}{arg\,max}

\newcommand{\myalg}{\texttt{Count-MORL}}

\newcommand{\B}{\fontseries{b}\selectfont}

\usepackage[capitalize,noabbrev]{cleveref}

\usepackage[textsize=tiny]{todonotes}

\icmltitlerunning{Model-based Offline Reinforcement Learning with Count-based Conservatism}

\begin{document}

\twocolumn[
\icmltitle{Model-based Offline Reinforcement Learning with Count-based Conservatism}

\icmlsetsymbol{equal}{*}

\begin{icmlauthorlist}
\icmlauthor{Byeongchan Kim}{SNU}
\icmlauthor{Min-hwan Oh}{SNU}
\end{icmlauthorlist}

\icmlaffiliation{SNU}{Seoul National University, Seoul, South Korea}
\icmlcorrespondingauthor{Min-hwan Oh}{minoh@snu.ac.kr}

\icmlkeywords{Offline Reinforcement Learning, Count-based, ICML}

\vskip 0.3in
]

\printAffiliationsAndNotice{}

\begin{abstract}
In this paper, we propose a model-based offline reinforcement learning method that integrates count-based conservatism, named {\myalg}. Our method utilizes the count estimates of state-action pairs to quantify model estimation error, marking the first algorithm of demonstrating the efficacy of count-based conservatism in model-based offline deep RL to the best of our knowledge.
For our proposed method, we first show that the estimation error is inversely proportional to the frequency of state-action pairs. Secondly, we demonstrate that the learned policy under the count-based conservative model offers near-optimality performance guarantees.
Through extensive numerical experiments, we validate that {\myalg} with hash code implementation significantly outperforms existing offline RL algorithms on the D4RL benchmark datasets.
The code is accessible at \href{https://github.com/oh-lab/Count-MORL}{https://github.com/oh-lab/Count-MORL}.
\end{abstract}

\section{Introduction}

Reinforcement Learning (RL) provides a paradigm in which an agent learns sequential actions within uncertain environments, all while aiming to maximize cumulative rewards. 
The empirical effectiveness of RL has been validated across diverse domains, underlining its capability to learn complex tasks~\citep{mnih2015human, silver2016mastering, silver2017mastering, fawzi2022discovering}. 
A defining feature of RL, which distinguishes it from other machine learning paradigms, is its inherent necessity for active information acquisition, in which the agent collects data through firsthand experimentation with the environment. This subfield of RL, characterized by direct interaction, is commonly referred to as \textit{online} RL.

Nevertheless, this approach with direct experimentation is often infeasible or potentially unsafe in many real-world applications, 
including but not limited to robotics~\cite{gu2017deep}, autonomous driving \cite{kiran2021deep}, and healthcare \cite{yu2021reinforcement}. 
Furthermore, even when direct intervention is viable, each interaction between the agent and the environment often incurs cost. 
Thus, if conditions permit, it would be advantageous for the agent to learn from readily available offline datasets, thereby negating the necessity for direct experimentation.

\textit{Offline} RL, also referred to as batch RL, is a subfield of the RL framework that relies solely on previously acquired static datasets. 
In this framework, the agent is given a collection of experiences that a behavior policy gathered and subsequently utilizes for policy learning without any additional interactions with the environment. 
Despite its advantages of learning from an offline dataset, 
offline RL also presents unique challenges~\cite{levine2020offline}.
Among these, striking a balance between enhancing generalization capabilities and circumventing undesired out-of-distribution (OOD) behaviors remains prominent.
The crux of finding such a balance pivots around managing \textit{distributional shift}. 
During policy evaluation, the application of Bellman updates to value functions involves querying the values of OOD state-action pairs. 
This can potentially lead to an accumulation of extrapolation errors. 
The issue becomes even more intricate due to the widespread use of high-capacity function approximators, such as neural networks, which further complicates its complexity.

To address this issue with OOD actions, various offline RL algorithms adapt conservatism in estimated values by inducing some form of pessimism. 
These methods include both model-free \citep{kumar2020conservative, shi2022pessimistic} and model-based \citep{yu2020mopo, kidambi2020morel, rafailov2021offline, lurevisiting, rigterrambo} approaches.
Recent model-based offline RL methods 
aim to harness conservative value estimates by using analytical performance bounds \citep{yu2020mopo, kidambi2020morel, rafailov2021offline}.
These methods adjust the estimated MDP model learned from the offline dataset to elicit conservative behavior by
imposing a penalty on the policy or rewards when the policy visits states where the estimated model's accuracy is likely low. 
If the learned policy takes actions in states where the accuracy of model prediction is high, it is plausible that the estimated value of the policy is likely to retain a high level of accuracy.
Specifically,
many existing model-based offline RL algorithms incorporate conservatism by estimating the model uncertainty and penalizing the reward function proportional to the estimated uncertainty~\citep{yu2020mopo, kidambi2020morel, rafailov2021offline}.

A recent work~\citep{lurevisiting} investigates a variety of estimated model uncertainty, showing that a 
different
choice of model uncertainty can significantly affect performances.
However, existing algorithms~\citep{yu2020mopo, rafailov2021offline} use the total variation distance between the estimated and true models as a penalty in theory, but such a distance is difficult to compute in practice.
Because of unreliable estimates of the model uncertainty, other approaches incorporate conservatism by regularizing the value function~\citep{yu2021combo} or modifying the estimated transition dynamics in an adversarial manner~\cite{rigterrambo} without model uncertainty quantification.
Hence, the questions of which choice of model uncertainty should be used and what conservatism is more suitable in practice remain open.

In this work, we propose a new method for model-based offline RL, Count-based Conservatism for Model-based Offline RL~(\texttt{Count-MORL}).
Our proposed method utilizes the estimated frequency (counts) of state-action pairs in the offline dataset to quantify the model estimation error.
The reward function is penalized by the model estimation error inversely proportional to 
the frequency of state-action pairs.
Using this count-based penalized reward, we construct the count-based conservative MDP model and learn a policy using this model.
However, when state and action spaces are not discrete, i.e., when state and action spaces are represented by high-dimensional  features or features that take continuous values,
it can be difficult to enumerate and quantify the exact counts of state-action pairs.
To efficiently implement our proposed method and address the aforementioned issue, we utilize hash code heuristics to approximate the frequency of state-action pairs~\citep{tang2017exploration}, previously used in online RL but has not been utilized for offline RL.
While we present our method with this hash code version of count estimation for clear and easy exposition, 
other count estimations~\cite{bellemare2016unifying, ostrovski2017count, fu2017ex2, martin2017count, machado2020count} can be utilized in our proposed method.
Another salient feature of our proposed method is the incorporation of uncertainty in count estimation, which is shown to affect the performance of the proposed method.
Hence, a suitable consideration of the count uncertainty can further enhance performance.
We show in our extensive numerical evaluations that our proposed count-based conservatism derives superior performances, consistently outperforming the existing methods in offline RL.

Our main contributions are summarized as follows:
\begin{itemize}
    \item We propose a new model-based offline RL method that incorporates count-based conservatism, named \texttt{Count-MORL}, where we utilize the count estimates of state-action pairs to quantify the model estimation error.
    To our best knowledge, this work is the first to show the efficacy of count-based conservatism in model-based offline deep RL.
    
    \item We provide two theoretical analyses established for our proposed method. First, the estimation error is bounded, inversely proportional to 
    the frequency of state-action pairs, which can be extended to consider count approximation, not just exact counts.
    Second, the learned policy under the count-based conservative model has a performance guarantee on near-optimality that depends on this estimation error and the count approximation.

    \item 
    The numerical experiments show that \texttt{Count-MORL} with hash code implementation significantly outperforms the existing offline RL algorithms in the D4RL benchmark datasets. The results support that count-based conservatism is both efficient and practical for model-based offline RL.
\end{itemize}

\section{Related Work}

Offline RL addresses the problem of learning policies from a logged static dataset.
Model-free offline algorithms do not require an estimated model and mainly belong to one of the three categories:
regularizing the learned policy to be close to the behavior policy~\citep{fujimoto2019off,wu2019behavior, liu2020provably, siegelkeep, fujimoto2021minimalist, kostrikovoffline},
quantifying the uncertainty with ensemble techniques to obtain a robust value function~\citep{agarwal2020optimistic, an2021uncertainty, kumar2019stabilizing},
and
adapting conservatism to a value function~\citep{kumar2020conservative, xie2021bellman}.
In contrast, model-based offline algorithms use an estimated model based on a fixed dataset to query model outputs for state transition and rewards so that algorithms can use them for planning to improve a policy.
However, training a policy
using an
inaccurately estimated model
can be harmful
\citep{janner2019trust}.
Such potential risk can be further exacerbated by distributional shifts.
Hence, it is important to \textit{correct} the insufficiently learned portion of estimated models.
Many existing model-based approaches~\citep{yu2020mopo, kidambi2020morel, matsushimadeployment, swazinna2021overcoming, yu2021combo, argensonmodel, lee2021representation, hishinuma2021weighted, rafailov2021offline, rigterrambo} address this challenge with various techniques,
which regularizes the value function without uncertainty estimation~\citep{yu2021combo, rigterrambo}.

A line of the model-based offline RL literature, most related to our paper, includes the works that utilize penalization on the estimated model~\citep{yu2020mopo, kidambi2020morel, rafailov2021offline}.
MOPO~\citep{yu2020mopo} penalizes the reward function with the estimation error between the true and estimated models based on the maximum aleatoric uncertainty.
MOReL~\citep{kidambi2020morel} constructs a pessimistic model using the unknown state-action detector based on a threshold of model uncertainty.
LOMPO~\citep{rafailov2021offline} handles image data with latent dynamics models and model uncertainty quantification based on the variance of log-likelihoods.
While new methods are being proposed for model-based offline RL, it is still unclear as to concretely what and how estimation error is used for practice.
Often, there appears to be a gap between what is proposed in theory and what is used in practice.
Our work aims to address this issue with a concrete uncertainty quantification using count estimation.

Count-based uncertainty quantification is widely used in online RL.
Count-based exploration methods~\cite{bellemare2016unifying, ostrovski2017count, fu2017ex2, martin2017count, tang2017exploration, machado2020count} are based on estimating the trial counts of state-action pairs and 
incorporating this count estimate into a bonus reward.
However, in offline RL, a count-based approach has been relatively under-utilized.
To our knowledge, the only work~\citep{hongconfidence}, which uses count-based conservatism for model-free offline RL, has focused on updating Q-functions by incorporating confidence levels derived from inverse state visitations.
Compared to~\citet{hongconfidence}, our proposed method has several distinct features.
First, in model-based offline deep RL, we utilize the count estimation of state-action pairs to quantify the model estimation error and construct the count-based conservatism model using this count estimation.
Second, unlike \citet{hongconfidence} which requires the last layer of the neural network of Q-function to be linear for approximating state visitations, our method can utilize various count-based methods used in online RL.
Additionally, for the first time, we address the issue of uncertainty in count estimation, which has been shown to enhance performance.

\section{Preliminaries}
\subsection{Markov Decision Process}
We consider a Markov decision process (MDP) specified by the tuples $M = \left( \gS,\gA,P,r, d_{0},\gamma \right)$, 
where $\gS$ is the state space, $\gA$ is the action space, $P:\gS \times \gA \to \Delta(\gS)$ is the transition dynamics, $r:\gS \times \gA \to \left[-R_{\mathrm{max}}, R_{\mathrm{max}}\right]$ is the reward function, $d_{0} \in \Delta(\gS)$ is the initial state distribution, and $\gamma \in \left[0,1\right)$ is the discount factor.
A policy $\pi:\gS \to \Delta(\gA)$ is a mapping from states to a probability distribution over actions.
The value function $V_{P,r}^{\pi}(s) := \mathbb{E}_{\pi, P} \left[ \sum_{t=0}^{\infty} \gamma^{t}r(s_{t},a_{t})|s_{0} = s \right]$ represents the expected cumulative discounted reward of $\pi$ under $P$ and $r$ when starting from state $s$.
We denote the discounted state visitation distribution of $\pi$ under $P$ using $d_{P}^{\pi}(s) := (1-\gamma) \sum_{t=0}^{\infty} \gamma^{t} d_{P}^{\pi}(s_{t}=s)$, where $d_{P}^{\pi} (s_{t}=s)$ is the probability of reaching state $s$ at a time-step $t$ under $\pi$ and $P$. 
Similarly, we denote the discounted state-action visitation distribution as $d_{P}^{\pi}\pare := d_{P}^{\pi}(s)\pi(a|s)$.

\subsection{Offline RL}
The goal of the RL agent in general is to obtain an optimal policy $\pi^{\ast}$ that maximizes the expected cumulative discounted reward under~$d_{0}$:
\begin{equation*}
    \max_{\pi} V_{M}^{\pi} := \mathbb{E}_{s \sim d_{0}}\left[V_{P,r}^{\pi}(s)\right] = \frac{1}{1-\gamma} \mathbb{E}_{\pare \sim d_{P}^{\pi}} \left[r\pare \right]
\end{equation*}
In offline RL, 
a policy is learned by 
utilizing a static dataset without additional interactions with the environment.
The agent is given an offline dataset $\gD = \left\{\left(s_{i}, a_{i}, r_{i}, s'_{i}\right) \right\}_{i=1}^{n}$ that consists of transition tuples from trajectories gathered in advance by some potentially unknown behavior policy~$\bp$.
Since $\gD$ is typically 
a (possibly very limited)
subset of the entire tuple space, finding the optimal policy using only this fixed dataset is not only challenging but also sometimes impossible especially when actions that belong to the optimal policy do not even exist in the given dataset.
Therefore, the goal of offline RL is to construct an algorithm that can learn 
a 
policy $\hat{\pi}$ minimizing the sub-optimality gap $V_{M}^{\pi^{\ast}} - V_{M}^{\hat{\pi}}$ 
computed based on
$\gD$ while hoping that $\hat{\pi}$ is close to $\pi^{\ast}$.

\subsection{Model-based Offline RL}
A model-based approach to offline RL utilizes an estimated transition dynamics model 
that approximates the true transition dynamics model.
Without loss of generality, we assume that the reward function $r(s,a)$ is known.\footnote{This is commonly assumed in the model-based RL literature~\citep{yang2019sample, yang2020reinforcement, zhou2021nearly, hwang2023model} without loss of generality since  learning $r(s,a)$ is considered much easier than learning $P(\cdot| s,a)$. 
If $r$ is unknown, it can be replaced with estimated reward $\lR$ learned together with~$\lP$.
}
Let $\lP(\cdot | s, a)$ denote the maximum likelihood estimator~(MLE) of 
the true state transition model
$P^{\star}(\cdot | s, a)$, computed based on the dataset $\gD_{s, a}:=\{(s_{i}, a_{i}, s'_{i})\}_{s_{i}=s, a_{i}=a}$ 
which is a subset of the entire offline dataset
$\gD$.
We denote the number of samples in $\gD_{s, a}$ as $\cnt := |\gD_{s, a}|$.
Then, we 
construct the estimated MDP $\lM := (\gS,\gA,\lP, r, d_{0}, \gamma)$ with the estimated transition model that approximates the true MDP $M^{\star} = (\gS,\gA,P^{*}, r, d_{0}, \gamma)$.
The existing methods in model-based offline RL use the estimated model $\lM$ to query synthetic trajectories by simulating $H$-step rollouts starting 
from a given state observed in the offline dataset~\citep{kidambi2020morel, yu2020mopo, yu2021combo, rafailov2021offline, rigterrambo}.
Similar to a replay buffer utilized in off-policy RL,
the synthetic trajectories are stored in a replay buffer $\gD_{\mathrm{model}}$.
However, an inaccurately estimated model for unobserved state-action pairs in $\gD$ can lead to poor performance of the learned policy~\citep{janner2019trust}.

To address this challenge, 
one of the most widely used approaches in model-based offline RL is to incorporate conservatism by estimating the model uncertainty and constructing an uncertainty-penalized MDP 
with a penalized reward where a penalty is proportional to the estimated uncertainty~\citep{yu2020mopo, kidambi2020morel, rafailov2021offline, lee2021representation}.
For example, an uncertainty-penalized MDP introduced in~\citet{yu2020mopo}
utilizes a penalized reward
$\tilde{r}\pare := r\pare-\lambda u\pare$, where $u\pare$ denotes an admissible error estimator for the state-action pair $(s,a)$.
Based on this uncertainty-penalized reward $\pR$, a learned policy maximizes the expected cumulative discounted rewards:
\begin{equation*}
    \mathbb{E}_{s \sim d_{0}}\left[V_{\lP,\pR}^{\pi}(s)\right] = \frac{1}{1-\gamma} \mathbb{E}_{\pare \sim d_{\lP}^{\pi}}\left[r\pare - \lambda u\pare \right].
\end{equation*}
However, one of the main issues in this approach \citep{yu2020mopo} as well as many other existing methods~\citep{kidambi2020morel, rafailov2021offline} is the 
unavailability of a readily usable uncertainty estimate
$u\pare$, which quantifies the estimation error between an estimated MDP and the unknown true MDP.

\section{Count-based Conservatism for Model-based Offline RL}
We propose a model-based algorithm for offline RL as described in Section~\ref{proposedalgorithm}.
The main idea of our proposed method is to compute the estimation error between the learned and true transition dynamics by quantifying the frequency of state-action pairs (or the frequency of the features of state-actions) in the offline dataset.
We use 
the estimated frequency of the observed data to construct a \textit{count-based} conservative MDP.

For our proposed method, we provide an estimation error bound based on the estimated frequency of state-action pairs in \cref{thm:estimationerror}~(in Section~\ref{theoreticalanalysis}).
For each state-action pair, we penalize the reward function, where the penalty is inversely proportional to the  estimated frequency.
With this penalized reward function, we construct a count-based conservative MDP.
In \cref{thm:sub-optimalitygap}, we 
provide a guarantee of the sub-optimality gap between the 
optimal policy under the true MDP and the learned policy under the conservative MDP.
In Section~\ref{proposedalgorithm},
we describe our algorithm that incorporates the aforementioned theoretical results to quantify the suitable conservatism and derives efficient learning in offline RL.

\subsection{Theoretical Analysis}
\label{theoreticalanalysis}
Suppose that we are given a hypothesis class of transition models
$\gM = \{ P : \gS \times \gA \rightarrow \Delta(\gS) \}$.
We assume realizability $P^{\star} \in \gM$, that is,  
the true transition model exists in the hypothesis class~\citep{agarwal2020flambe, ueharapessimistic}.
We denote a total variation distance between two distributions $P_{1}$ and $P_{2}$ as $\mathrm{TV}(P_{1}, P_{2})$.
Now, we begin by presenting the following theorem that establishes an upper bound of the total variation distance between the learned and true transition dynamics.
The theorem is adapted from Theorem~21 in~\citet{agarwal2020flambe}.

\begin{theorem}[Estimation error of transition dynamics]
    \label{thm:estimationerror}
    Fix $\delta \in (0,1)$, assume $|\gM| < \infty$ and $P^{\star} \in \gM$.
    Given a state-action pair $\pare$ is observed in $\gD$ with $\gD_{s, a} =\{(s_{i}, a_{i}, s'_{i})\}_{s_{i} = s, a_{i} = a}$ and $\cnt = |\gD_{s, a}|$. 
    Define the MLE of transition dynamics as 
    \begin{equation*}
        \lP(\cdot \mid s, a) \in \argmax_{P \in \gM} \sum_{(s,a,s') \in \gD_{s, a}} \log P(s' \mid s, a)
    \end{equation*}
    for a given $\pare$.
    Then with probability at least $1-\delta$,
    \begin{equation*}
         \mathrm{TV}\!\left(\lP(\cdot \mid s,a), P^{\star}(\cdot \mid s,a) \right) \le \sqrt{\frac{2\log(|\gM|/\delta)}{\cnt}}.
    \end{equation*}
\end{theorem}

\textbf{Discussion of \cref{thm:estimationerror}.}
\cref{thm:estimationerror} states that the estimation error between the estimated and true transition dynamics is smaller for the state-action pairs that are more frequently observed in the offline dataset $\gD$, and provides the upper bound on the rate of the convergence in terms of the frequency, i.e., $O(1/\sqrt{n(s,a)})$.
Such quantification of the model estimation error is critical for offline RL algorithms since the penalization on reward functions needed for learning conservative policies depends on the specification of the error bound.
The key of this theorem is that the upper bound of the model estimation error can be determined based on the number of observations in the offline dataset, which we aim to estimate in the proposed algorithm (particularly when the environment is not tabular).

As an immediate corollary, we expand the estimation error for individual state-action pairs from the offline dataset $\gD$ to the entire state-action space.

\begin{corollary} 
    \label{cor:errorbound}
    Given a state-action pair $\pare \in \gS \times \gA$,
    with probability at least $1-\delta$, the estimated transition dynamics $\lP$ satisfies the following inequality by \cref{thm:estimationerror}:
    \begin{align}
        \mathrm{TV}\!\left(\lP(\cdot \mid s,a), P^{\star}(\cdot \mid s,a)\right) \le
        C_{\lP}^{\delta}\pare , \nonumber
    \end{align}
    where $C_{\lP}^{\delta}\pare := \min \left(1, \sqrt{\frac{2\log(|\gM|/\delta)}{\cnt}} \right)$.
\end{corollary}
We define the estimation error bound based on the true count for the estimated transition dynamics $\lP$ as $C_{\lP}^{\delta}:\gS \times \gA \rightarrow \left[0, 1\right]$.
By the definition of total variation distance, the estimation error for all state-action pairs is bounded by~$1$.
There are two cases in which the estimation error takes the value of $1$.
First, for unobserved state-action pairs in $\gD$, we cannot estimate the transition dynamics or compute the estimation error.
Second, when $\cnt$ is less than $\sqrt{2\log(|\gM|/\delta)}$, the estimation error becomes greater than~$1$.
Excluding these cases, all estimation errors are inversely proportional to observation counts $\cnt$.

So far, we have assumed that we can compute exact counts for each observation $\cnt$ which is only applicable to tabular settings. 
However, for large state and action spaces, or continuous state-action pairs, 
computing the exact frequency of state-action pairs may be intractable. 
To this end, we introduce the notion of \textit{approximate counts} $\hat{n}$ to approximate the true counts.

\begin{definition}[Estimation error with approximate count]
Define the approximate count as $\hat{n}:\gS \times \gA \rightarrow \mathbb{R}$, which approximates the true count $n$, and the estimation error bound based on the approximate count as $\widehat{C}_{\lP}^{\delta}:\gS \times \gA \rightarrow \left[0, 1\right]$:
\begin{equation*}
    \widehat{C}_{\lP}^{\delta}\pare := \min \left(1, \sqrt{\frac{2\log(|\gM|/\delta)}{\hcnt}} \right).
\end{equation*}
\end{definition}

We allow an approximation error incurred by $\hat{n}$. We define the maximal approximation error as follows.
\begin{definition}[Approximation error of counts]
Define the maximal approximation error between estimation error bounds based on the true count and approximate count over all state-action pairs as
\begin{equation*}
    \epsilon := \sup_{(s,a) \in \gS \times \gA} \left| C_{\lP}^{\delta}\pare - \widehat{C}_{\lP}^{\delta}\pare \right|.
\end{equation*}    
\end{definition}
The following lemma shows the gap of returns between the estimated MDP $\lM$ and the true MDP $M^{\star}$ of any given policy~$\pi$. 
This result is an adaptation of Lemma~$4.1$ based on the telescoping lemma in \citet{yu2020mopo} with the approximate count.

\begin{lemma}[Value gap of estimated model]
    \label{lemma:valuegap}
    Suppose $M^{\star}$ and $\lM$ be two MDPs with the true transition dynamics $P^{\star}$ and estimated transition dynamics $\lP$, respectively.
    Given the estimation error bound based on the approximate count $\widehat{C}_{\lP}^{\delta}$ and
    the maximal approximation error $\epsilon$.
    Then, with probability at least $1-\delta$, for any policy $\pi$,
    \begin{equation*}
        V_{\lM}^{\pi} - V_{M^{\star}}^{\pi}
        \le
        \frac{\gamma R_{\mathrm{max}}}{(1-\gamma)^{2}} \mathbb{E}_{\pare \sim d_{\lP}^{\pi}}\! \left[ \widehat{C}_{\lP}^{\delta}\pare \right]
        + \frac{\gamma R_{\mathrm{max}}}{(1-\gamma)^{2}} \epsilon \,.
    \end{equation*}
\end{lemma}
\textbf{Discussion of Lemma~\ref{lemma:valuegap}.}
\cref{lemma:valuegap} states that the value gap between the estimated and true model under any policy~$\pi$ 
is bounded by
the model estimation error under the visitation distribution $d_{\lP}^{\pi}$
and the count approximation error.
Our approach of quantifying the model estimation error using the approximate count $\hcnt$ for all state-action pairs, 
instead of an admissible error estimator $u(s,a)$, 
provides a practical way of constructing a reward penalty concretely defined once an approximate count is provided.

Note that $\widehat{C}_{\lP}^{\delta}$ is a function of $\lP$ and $\pi$.
In order to understand the implication of Lemma~\ref{lemma:valuegap}, 
first, consider 
a special
case when $\pi = \bp$.
Then, the policy takes actions
frequently observed
in $\gD$ and visits states frequently observed in $\gD$,
so that
the estimation error bound for state-action pairs will be generally small by~\cref{thm:estimationerror}.
Hence, the expectation of $\widehat{C}_{\lP}^{\delta}$ under $d_{\lP}^{\bp}$ will be small.
Conversely, if a policy 
takes actions (or visits states) less observed or unobserved 
in $\gD$, then 
the estimation error bound for unobserved state-action pairs is likely to be large.
Thus, the expectation of $\widehat{C}_{\lP}^{\delta}$ under $d_{\lP}^{\pi}$ tends to be 
large.
Note that the maximal approximation error $\epsilon$ does not depend on $\lP$ or $\pi$, that is, even if the estimation of the model is accurate the error incurred by the approximation still exists.
Hence, when the approximation error is sufficiently small,
the value function will be efficiently learnable.

The key insight of \cref{lemma:valuegap} is that 
the value gap for the learned policy, which maximizes 
the value function
with respect to $\lM$, 
can be large 
when the learned policy takes actions (or visits states) beyond the offline dataset $\gD$, or when the count approximation is inaccurate. 
Using this insight, our proposed method augments count-based conservatism by incorporating the reward penalty based on count-based estimation error.
Also, our method even addresses the uncertainty in count approximation.

Now, we define a count-based conservative MDP as follows.

\begin{definition}[Count-based conservative MDP]\label{def:conservative-MDP}
    Define the count-based conservative MDP $\pM := (\gS, \gA, \lP, \pR, d_{0}, \gamma)$ with the estimated transition dynamics $\lP$ and count-based penalized reward $\pR\pare := r\pare - \frac{\gamma R_{\mathrm{max}}}{1-\gamma} \widehat{C}_{\lP}^{\delta}\pare$.
\end{definition}

We invoke the following corollary to demonstrate that $\pM$ is conservative in that the value of a policy under $\pM$ is generally not higher than its value under $M^\star$.
\begin{corollary}
    \label{cor:conservatismMDP}
    Given the true MDP $M^{\star}$ and the count-based conservative MDP $\pM = (\gS, \gA, \lP, \pR, d_{0}, \gamma)$ as defined in Definition~\ref{def:conservative-MDP}.
    Then, with probability at least $1-\delta$, for any policy~$\pi$,
    \begin{equation*}
        V_{M^{\star}}^{\pi} \ge V_{\pM}^{\pi} - \frac{\gamma R_{\mathrm{max}}}{(1-\gamma)^{2}}\epsilon \,.
    \end{equation*}
\end{corollary}
Here, the conservatism error is given by a multiple of the maximal approximation error $\epsilon$. That is, if $\epsilon$ is $0$, then the value under $\pM$ is pessimistic with high probability.

We define the learned policy $\hat{\pi}$ that maximizes the value function with respect to $\pM$, that is, $\hat{\pi}:= \argmax_{\pi}V_{\pM}^{\pi}$.
Then, our main theorem (Theorem~\ref{thm:sub-optimalitygap}) provides a performance guarantee on the learned policy $\hat{\pi}$ under $M^{\star}$.
\begin{theorem}[Sub-optimality gap]
    \label{thm:sub-optimalitygap}
    Given the estimation error bound $\widehat{C}_{\lP}^{\delta}$ based on the approximate count $\hat{n}$ and
    the maximal count approximation error $\epsilon$,
    with probability at least $1 - \delta$,
    the learned policy~$\hat{\pi}$ under the count-based conservative MDP $\pM$ satisfies
    \begin{align*}
        V_{M^{\star}}^{\hat{\pi}} 
        &\ge
        \sup_{\pi} \left\{V_{M^{\star}}^{\pi} - \frac{2\gamma R_{\mathrm{max}}}{(1-\gamma)^{2}}\mathbb{E}_{\pare \sim d_{\lP}^{\pi}} \left[ \widehat{C}_{\lP}^{\delta}\pare \right] \right\} \nonumber \\
        &\quad - \frac{2\gamma R_{\mathrm{max}}}{(1-\gamma)^{2}}\epsilon \,.
    \end{align*}
    In particular, for the optimal policy $\pi^{\ast}$,
    \begin{align*}
        V_{M^{\star}}^{\pi^{\ast}} - V_{M^{\star}}^{\hat{\pi}} 
        &\le 
        \frac{2\gamma R_{\mathrm{max}}}{(1-\gamma)^{2}}\mathbb{E}_{\pare \sim d_{\lP}^{\pi^{\ast}}} \left[ \widehat{C}_{\lP}^{\delta}\pare \right] \nonumber \\
        &\quad + \frac{2\gamma R_{\mathrm{max}}}{(1-\gamma)^{2}}\epsilon \,.
    \end{align*}
\end{theorem}
\textbf{Discussion of \cref{thm:sub-optimalitygap}.}
\cref{thm:sub-optimalitygap} establishes that the 
value
of the learned policy $\hat{\pi}$ under the true MDP $M^{\star}$ can be controlled in terms of the trade-off between the value under $M^{\star}$ and the expected estimation error  under $\lM$ and 
the count approximation error.
\cref{thm:sub-optimalitygap} has two important interpretations.
First, the learned policy $\hat{\pi}$ will perform as least as well as the behavior policy $\bp$ under $M^{\star}$ since the expectation of $\widehat{C}_{\lP}^{\delta}$ under $d_{\lP}^{\bp}$ is small.
Second, $\hat{\pi}$ 
aims to find the optimal
balance between the value under $M^{\star}$ 
and the expectation of $\widehat{C}_{\lP}^{\delta}$ under $d_{\lP}^{\pi}$.
There exists a tradeoff between the two quantities since
a policy that increases $V_{M^{\star}}^{\pi}$ 
may also increase the expectation of $\widehat{C}_{\lP}^{\delta}$ under $d_{\lP}^{\pi}$.
A learned policy different
from $\bp$ 
may take actions not observed in the dataset or visit unobserved states with high-value functions.
Then, the resulting $\widehat{C}_{\lP}^{\delta}\pare$ for those state-action pairs will be large since $\hcnt$ is small.
Hence, this conservative MDP encourages the agent to find a near-optimal policy 
but simultaneously discourages the agent to take actions that are unobserved in $\gD$.
The second inequality in \cref{thm:sub-optimalitygap} shows that the sub-optimality gap between the optimal policy $\pi^{\ast}$ and the learned policy $\hat{\pi}$ under $M^{\star}$ depends on the expectation of $\widehat{C}_{\lP}^{\delta}$ under $d_{\lP}^{\pi^{\ast}}$.
If the quality of the offline dataset $\gD$ is high --- that is, if the behavior policy $\bp$ is similar to $\pi^{\ast}$, then the expectation of $\widehat{C}_{\lP}^{\delta}$ 
becomes small, allowing $\hat{\pi}$ to be closer to $\pi^{\ast}$.

\subsection{Proposed Algorithm: \myalg}
\label{proposedalgorithm}

We present our algorithm,
\textit{Count-based Conservatism for Model-based Offline RL} (\myalg).
The algorithm is summarized in Algorithm~\ref{alg:count-MORL}.
Our algorithm learns the estimated transition model and computes the approximate counts of state-action pairs in the offline dataset to penalize the reward function, so as to induce count-based conservatism.
Our proposed algorithm is generic in terms of incorporating count estimation. Hence, one can utilize any type of count approximation studied in the online RL literature~~\citep{bellemare2016unifying, tang2017exploration, martin2017count, ostrovski2017count, machado2020count}.

Another salient feature of our method is 
incorporation of the uncertainty of the count estimation, shown in Algorithm~\ref{algo:count_est} which is a sub-routine of \myalg. Algorithm~\ref{algo:count_est} allows three possible types of count estimation: lower confidence~(\texttt{LC}), average~(\texttt{AVG}), and upper confidence~(\texttt{UC}). As shown in the numerical experiments of Section~\ref{sec:experiments}, a certain type of count estimation performs superior to others depending on how samples in the offline dataset were collected (see Table~\ref{table:countestimation} for more details).
Hence, the consideration of uncertainty in count estimation is crucial.

\begin{algorithm}[t]
    \caption{: Count-based Conservatism for Model-based Offline RL (\myalg)}
    \label{alg:count-MORL}
    \begin{algorithmic}[1]
        \REQUIRE counting functions \{$n_{i}:\mathbb{R}^{d} \rightarrow \mathbb{R}^{+} \cup \{0\} \}_{i=1}^{N}$, reward penalty coefficient~$\beta$, rollout horizon~$H$, rollout batch size~$B$, offline dataset $\gD$
        \STATE Train an ensemble of $N$ dynamics models $\{ \lP_{i} \}_{i=1}^{N}$ and feature mappings $\{ \phi_{i} : \gS \times \gA \rightarrow \mathbb{R}^{d}\}_{i=1}^{N}$ on $\gD$.
        \STATE Initialize policy $\pi$ and the replay buffer $\gD_{\mathrm{model}} \leftarrow \varnothing$.
        \FOR{\texttt{epoch} = $1, 2, \cdots$}
            \FOR{\texttt{rollout} = $1, 2, \cdots, B$}
                \STATE Sample initial rollout state $s_{1}$ from $\gD$.
                \FOR{$t = 1, 2, \cdots, H$}
                    \STATE Sample an action $\hat{a}_{t} \sim \pi(\hat{s}_{t})$
                    \STATE Sample a dynamics model $\lP$ from $\{\lP_{i}\}_{i=1}^{N}$
                    \STATE Sample $\hat{s}_{t+1}, \hat{r}_{t} \sim \lP(\hat{s}_{t}, \hat{a}_{t})$
                    \STATE Compute $\hat{n}(\hat{s}_{t}, \hat{a}_{t})$ using Algorithm~\ref{algo:count_est}
                    \STATE Compute \\ \hfil $\pR_{t} = 
                        \begin{cases}
                            \hat{r}_{t} - \frac{\beta}{\sqrt{\hat{n}(\hat{s}_{t},a_{t})}} & \texttt{if }\hat{n}(\hat{s}_{t}, \hat{a}_{t}) > 0\\
                            \hat{r}_{t} - \beta                                     & \texttt{otherwise}.
                        \end{cases}$
                    \STATE Add sample $(\hat{s}_{t}, \hat{a}_{t}, \pR_{t}, \hat{s}_{t+1})$ to $\gD_{\mathrm{model}}$.
                \ENDFOR
            \ENDFOR
            \STATE Draw samples from $\gD \cup \gD_{\mathrm{model}}$ to update $\pi$.
        \ENDFOR
    \end{algorithmic}
\end{algorithm}
\begin{algorithm}[t]
    \caption{: Count Estimation}
    \label{algo:count_est}
    \begin{algorithmic}[1]
        \REQUIRE standard deviation coefficient~$\alpha$, feature mappings~$\{ \phi_{i} \}_{i=1}^{N}$, counting functions~\{$n_{i}\}_{i=1}^{N}$
        \FOR{$i=1, 2, \cdots, N$}
            \STATE Compute $\phi_{i}\pare$
            \STATE Compute  $\hat{n}_{i}\pare = n_{i}(\phi_{i}\pare)$
        \ENDFOR
        \STATE Compute $\bar{n}\pare = \frac{1}{N}\sum_{i=1}^{N} \hat{n}_{i}\pare$
        \STATE Compute $\sigma\pare = \sqrt{\frac{\sum_{i=1}^{N} \left\{\hat{n}_{i}\pare - \bar{n}\pare \right\}^{2}}{N-1}}$
        \STATE Compute \\ \hfil $\hat{n}(s, a) = 
                        \begin{cases}
                            \bar{n}\pare - \alpha \sigma\pare & \texttt{if}\text{ \texttt{LC} count} \\
                            \bar{n}\pare                      & \texttt{if}\text{ \texttt{AVG} count}\\
                            \bar{n}\pare + \alpha \sigma\pare & \texttt{if}\text{ \texttt{UC} count}.
                        \end{cases}$
        
        \STATE \textbf{return} an approximate count $\hat{n}\pare$
    \end{algorithmic}
\end{algorithm}

Algorithm~\ref{alg:count-MORL} presents a general version of our proposed method with feature mappings.\footnote{For tabular settings, feature representation for each state-action pair can be given by one-hot encoding.}
The first step of our algorithm is to train an ensemble of $N$ dynamics models $\{\lP_{i}(s', r|s, a)\}_{i=1}^{N}$ and feature mappings $\{\phi_{i}\pare\}_{i=1}^{N}$.
Given a state-action pair, each composition function of the feature mapping and counting function, \{$n_{i}(\phi_{i}\pare)\}_{i=1}^{N}$, are obtained.
A mean and a standard deviation are calculated by using these composition functions as $\bar{n}\pare$ and $\sigma\pare$.
When we have prior information available about the offline dataset, the count estimation (\texttt{LC}, \texttt{AVG}, or \texttt{UC}) can be chosen appropriately.
For example, 
if the offline dataset 
is given by a clearly suboptimal behavior policy 
(in other words, the observed trajectories significantly differ from trajectories possibly given by an optimal policy),
then it is sensible to consider a higher level of pessimism on the observed samples, therefore utilizing \texttt{LC} counts in this case.
Then, a learned policy is more inclined to take actions (and visit states) outside the offline dataset --- since the penalization on observed samples is increased while the penalization on unobserved data is fixed.
Conversely, if the offline dataset is given by a behavior policy which is near-optimal, then it is appropriate to utilize \texttt{UC} counts to penalize with a lower level of pessimism on the observed samples.
In that case, a learned policy is encouraged to take actions within the offline dataset.

\subsection{Architecture and Efficient Implementation}\label{sec:architecture_implementation}

\begin{figure}
    \centering
    \includegraphics[height=4.25cm,trim={2.8cm 11cm 15cm 0.5cm},clip]{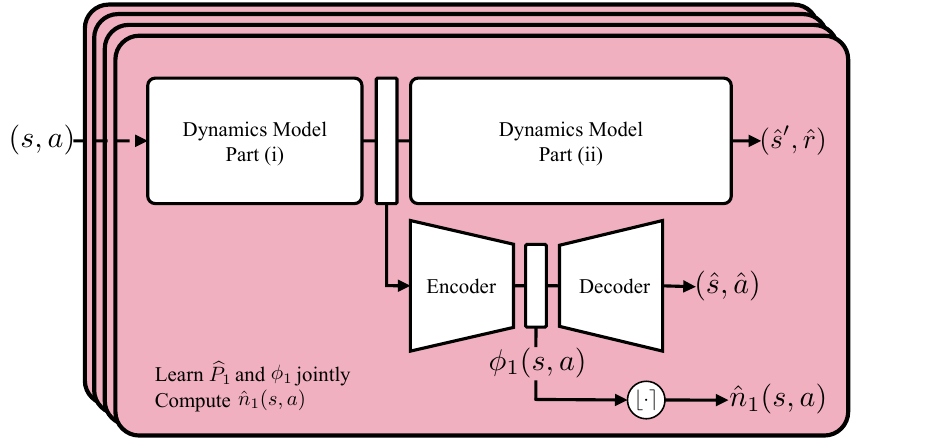}
    \caption{Efficient implementation of \texttt{Count-MORL} with hash codes: given a state action pair as input, for each model, taking the feature mapping $\phi$ and computing the counting function $\hat{n}$ using $\phi$.}
    \label{fig:model}
\end{figure}

Although {\myalg} is a comprehensive method not limited to a specific architecture, we provide a concrete architectural example (used for the experiments in Section~\ref{sec:experiments}) that details its efficient implementation using \textit{hash-code} count estimation, adapted from a technique used in the online RL literature \citep{tang2017exploration}. 
The overall architecture is depicted in Figure~\ref{fig:model}.

The estimated dynamics model for continuous state transitions is constructed using a neural network that predicts a Gaussian distribution over the estimated next state and reward, similar to the approach of \citet{yu2020mopo}. 
This model is linked to an autoencoder at the hidden layer. 
The autoencoder's input is the output of this hidden layer, which includes a bottleneck layer comprised of $d$ sigmoid functions.
By rounding the sigmoid activations of the bottleneck layer to their nearest binary values, we convert all state-action pairs into binarized $d$-dimensional binary vectors. 
As the learning process advances, the latent representation of these state-action pairs, along with their corresponding binarized $d$-dimensional vectors, become more stable.

To compute the frequency of potentially high-dimensional or continuous state-action pairs, we count each unique binary vector corresponding to each state-action pair.
However, since the binary vector corresponding to each state-action pair can vary based on different estimated models (given that we use an ensemble of transition models), 
we apply an approximate count computed from the count obtained for each learned model, as outlined in Algorithm~\ref{algo:count_est}.

\section{Experiments}\label{sec:experiments}

In this section, we present our experimental procedures and their respective outcomes, seeking to answer the following pertinent questions:

\begin{enumerate}[(i)]
    \item
    Can the proposed method produce
    approximate counts that accurately estimate the true counts for each state-action pair?

    \item
    How does the performance of \texttt{Count-MORL} measure up against the previous state-of-the-art (SOTA) benchmarks in offline RL?

    \item
    How do count estimation methods, which account for uncertainty in count estimation, perform when exposed to different types of datasets?
\end{enumerate}

\subsection{Experimental Setup}

\begin{figure}
    \centering
    \begin{subfigure}[]{0.3\linewidth}
        \centering
        \includegraphics[width=2.5cm,trim={0 2cm 17cm 0},clip]{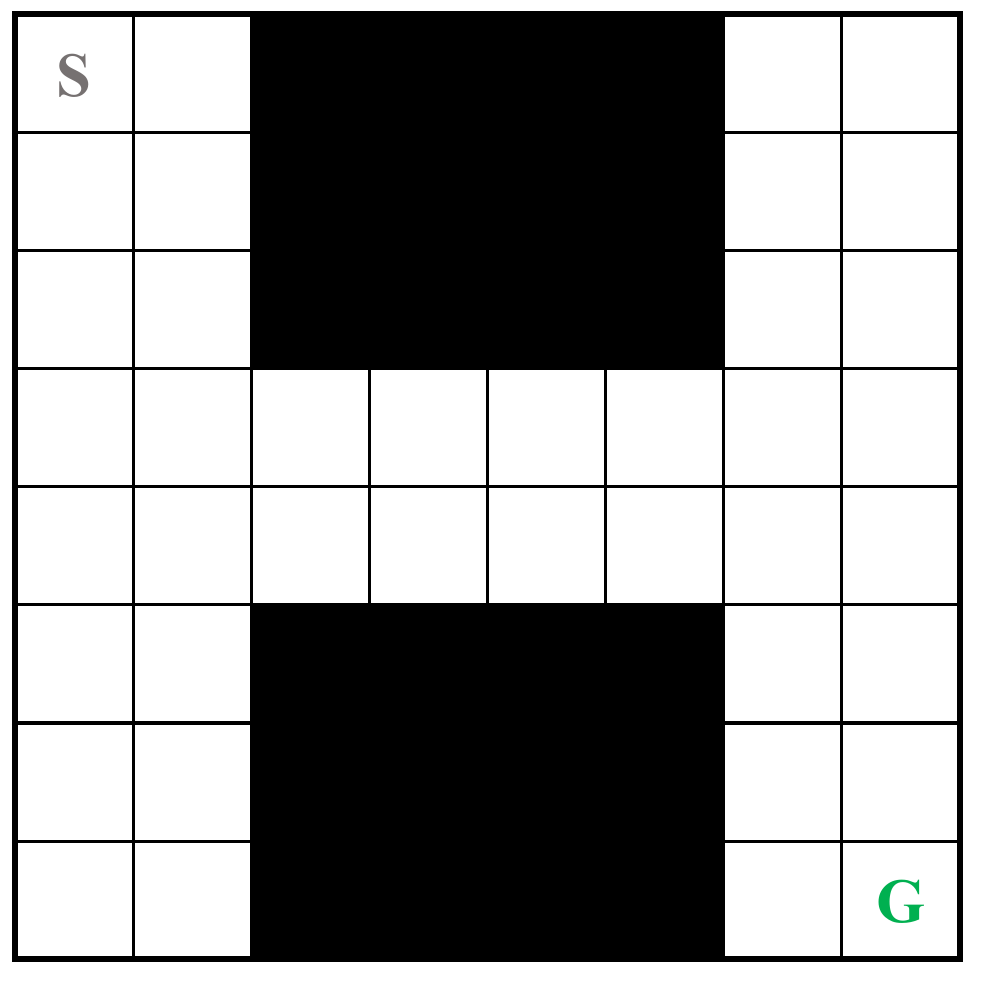}
        \caption{Bridge}
    \end{subfigure}
    \hfill
    \begin{subfigure}[]{0.3\linewidth}
        \centering
        \includegraphics[width=2.5cm,trim={0 2cm 17cm 0},clip]{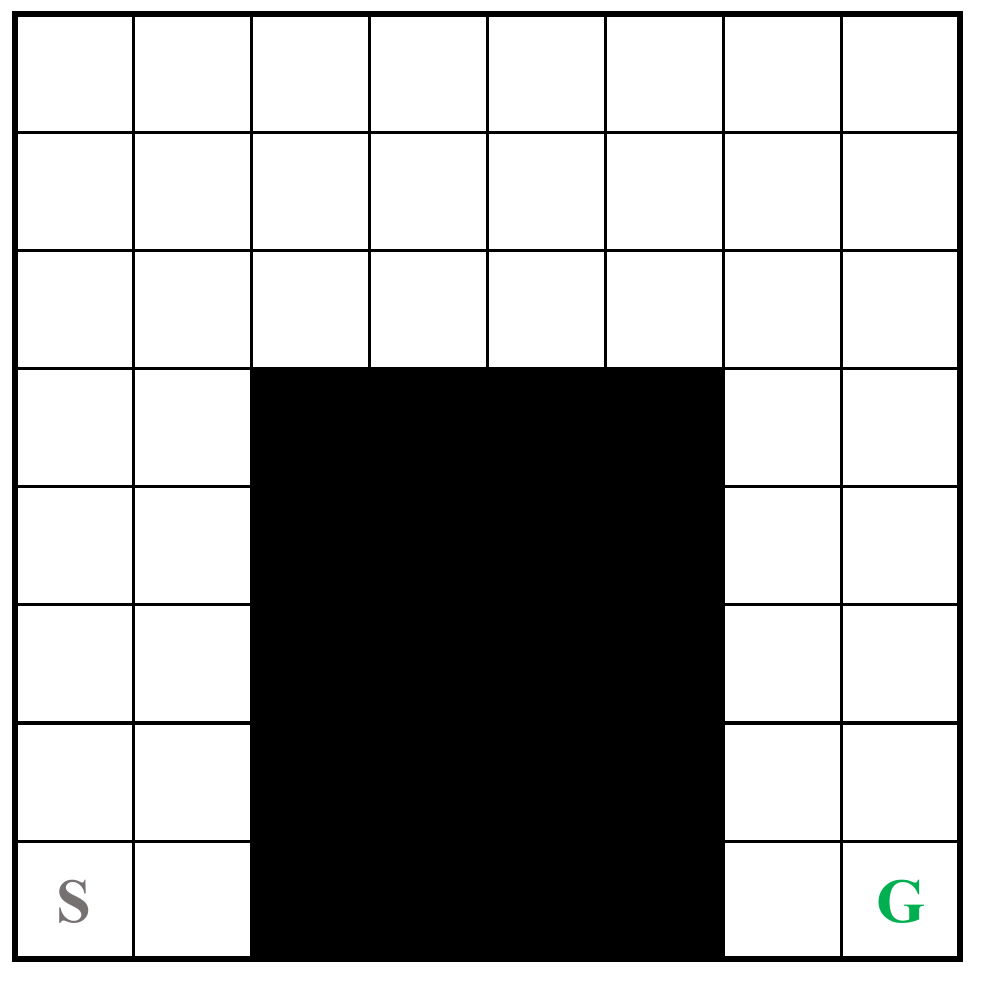}
        \caption{Cliff}
    \end{subfigure}
    \hfill
    \begin{subfigure}[]{0.3\linewidth}
        \centering
        \includegraphics[width=2.5cm,trim={0 2cm 17cm 0},clip]{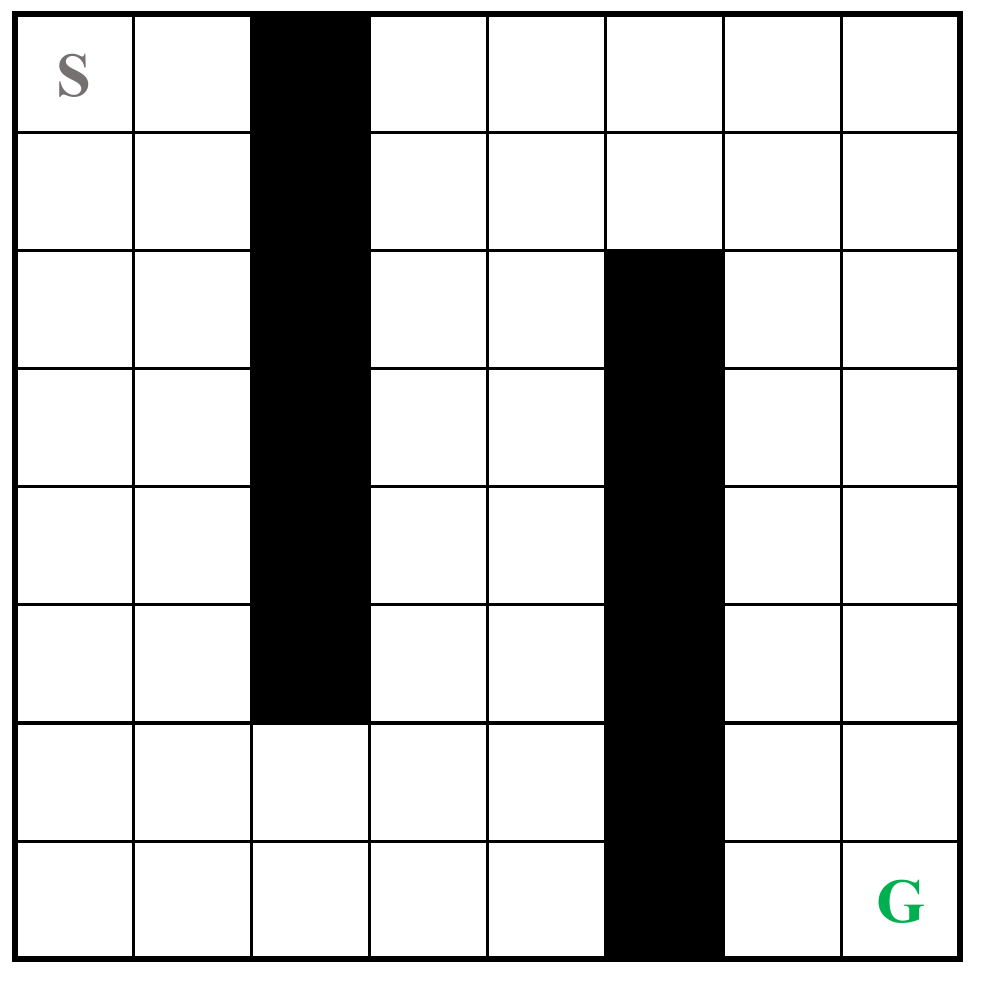}
        \caption{Zigzag}
    \end{subfigure}
    \caption{3 types of 8$\times$8 Grid World, Bridge, Cliff, and Zigzag. The objective is to reach a fixed goal state (\textcolor[rgb]{0, 0.69, 0.3}{G}) from a fixed start state (\textcolor[rgb]{0.45, 0.44, 0.44}{S}) while avoiding the lava area colored in black.}
    \label{figure:gridworld}
\end{figure}

\textbf{Grid-World.}
To address the question (i), we start by precisely counting the number of samples for each state-action pair in the Grid-World setting. As we have the true count for each state-action pair at hand, we can evaluate the discrepancy between the approximate counts and the authentic counts.
We employ four types of Grid-World environments ---\textit{Empty}, \textit{Bridge}, \textit{Cliff}, and \textit{ZigZag} --- as depicted in Figure~\ref{figure:gridworld} \citep{biedenkapp2021temporl}. 
In each of these environments, we amass transition samples from a replay buffer of the policy trained via Q-learning.
The \textit{Empty} dataset contains $10^{6}$ samples and covers all state-action pairs. In contrast, the \textit{Bridge}, \textit{Cliff}, and \textit{ZigZag} datasets encompass $1.6\times 10^{5}$, $1.4 \times 10^{5}$ and $3 \times 10^{5}$ transition samples, respectively, but they do not include samples for all actions taken within the lava zones, represented by black grids.

\textbf{MuJoCo.}
We evaluate \texttt{Count-MORL} on datasets in the D4RL benchmark~\citep{fu2020d4rl}, which comprises a total of $12$ datasets from $3$ different environments (\textit{HalfCheetah}, \textit{Hopper}, and  \textit{Walker2d}), each with $4$ dataset types (\textit{Random, Medium, Medium-Replay, Medium-Expert}).
\textit{Random} dataset contains $10^{6}$ transition samples collected by a random policy.
\textit{Medium} dataset contains $10^{6}$ transition samples collected by a partially trained SAC policy.
\textit{Medium-Replay} dataset contains $10^{5}$ ($2 \times 10^{5}$ for \textit{Walker2d}) transition samples, which are from the replay buffer that accumulates samples obtained by interacting with the environment until the policy trained using SAC reaches the performance of \textit{Medium} agent.
\textit{Medium-Expert} dataset contains $10^{6}$ transition samples, a mixture of sub-optimal and expert samples obtained by the partial and fully-trained policy.

\textbf{Counting Method.}
As explained earlier, there are three count estimation methods: \texttt{LC}, \texttt{AVG}, and \texttt{UC} counts.
When using \texttt{LC} and \texttt{UC} count methods, for each dataset, we choose the standard deviation coefficient $\alpha$ as $0.5$.
By changing the value of $\alpha$, the approximate count can be adjusted to be either large or small (Line 7 in Algorithm~\ref{algo:count_est}).

\textbf{Hyperparameter Details.}
Our algorithm adopts its foundational hyperparameters from the MOPO framework~\citep{yu2020mopo}.
While MOPO typically utilizes relatively short rollout lengths, such as 2 or 5 steps, recent research~\citep{lurevisiting} emphasizes the crucial role that the rollout length parameter, denoted by $h$, plays in the performance of model-based offline RL algorithms. Therefore, we extend the rollout length to accommodate up to 20 steps. We select an optimal rollout length and a reward penalty coefficient from the following potential values: $h \in \{5, 20\}$ and $\beta \in \{0.5, 1, 3, 5\}$.

\textbf{Hyperparameter Details of Autoencoder.}
We employ a $d$-dimensional binary vector to count the samples within our offline dataset. Our experiments explore the performance with different dimensions of this binary vector, specifically considering five distinct values: $d \in \{16, 32, 50, 64, 80\}$.

\textbf{Baselines.}
We compare our algorithm against the state-of-the-art model-based algorithms, RepBSDE \citep{lee2021representation}, RAMBO \citep{rigterrambo}, COMBO \citep{yu2021combo}, MOPO \citep{yu2020mopo} and MOReL \citep{kidambi2020morel}, and model-free algorithms, CQL \citep{kumar2020conservative} and TD3+BC \citep{fujimoto2021minimalist}, for offline RL.
COMBO, MOPO, and CQL evaluate the performance of their algorithms on the MuJoCo-v0 datasets.
However, RepB-SDE, RAMBO, MOReL, and TD3+BC papers include their performance on the MuJoCo-v2 datasets.
We provide normalized scores for \texttt{Count-MORL} and reproduce the performance of COMBO, MOPO, and CQL algorithms on the MuJoCo-v2 datasets.
This allows for a fair comparison with all SOTA offline RL algorithms, which have also been evaluated on the MuJoCo-v2 datasets.

\subsection{Results}
\subsubsection{Results on Grid-World}

We assess the accuracy of our approach by comparing the approximate counts with the true counts in the 8$\times$8 Grid-World environments. The autoencoder, which is connected to the dynamics model, uses the output from the dynamics model's hidden layer as its input, as detailed in our neural network architecture (see Figure~\ref{fig:model}). Despite the intricacies of this setup, we find that the approximate count, derived from a hash code, matches the true count for each state-action pair across all the Grid-World environments that we tested, \textit{Bridge}, \textit{Cliff}, and \textit{Zigzag}. These findings underscore the accuracy and robustness of our count estimation method. More detailed results are provided in Appendix~\ref{appendix:grid-world}.

\subsubsection{Results on D4RL tasks}
\begin{table*}[t]
\caption{Results for D4RL datasets. Each number is the normalized score proposed in \citealt{fu2020d4rl} of the policy during the last $5$ iterations averaged over $5$ seeds,
where $\pm$ denotes the standard deviation over seeds. 
We take the results of RepB-SDE, RAMBO, MOReL, and TD3+BC from their papers. We reproduce the results of COMBO, MOPO, and CQL with MuJoCo-v2 datasets. 
We include the score of behavior cloning~(BC) for comparison. Bold numbers are the scores within 2\% of the highest score in each environment.}
\begin{center}
\begin{small}
\sisetup{separate-uncertainty}
\begin{adjustbox}{max width=\textwidth}
\begin{tabular}{
c
l|
S[detect-weight, table-format = 3.1(2)]|
S[detect-weight, table-format = 2.1]
S[detect-weight, table-format = 2.1]
S[detect-weight, table-format = 3.1(3)]
S[detect-weight, table-format = 2.1(3)]
S[detect-weight, table-format = 3.1]|
S[detect-weight, table-format = 3.1(3)]
S[detect-weight, table-format = 3.1]|
S[detect-weight, table-format = 3.1]
}
\toprule
 &  & {} & \multicolumn{5}{c|}{Model-based baselines} & \multicolumn{2}{c|}{Model-free baselines} &  \\
{Dataset type} & {Environment} & {\myalg} & {RepB-SDE} & {RAMBO} & {COMBO} & {MOPO} & {MOReL} & {CQL} & {TD3+BC} & {BC} \\
\midrule
\multirow{3}{*}{Random}         &   Halfcheetah    & \B 41.0(09)   & 32.9 & 40.0    & 36.7(21)   & 34.0(28)   & 25.6      & 26.6(08)     & 11.0     & 2.1   \\
                                &   Hopper         & 30.7(13)      & 8.6  & 21.6    & 7.8(09)    & 7.0(19)    & \B 53.6   & 9.4(06)      & 8.5      & 9.8   \\
                                &   Walker2d       & 21.9(02)      & 21.1 & 11.5    & 5.9(03)     & 5.6(57)    & \B 37.3   & -0.4(08)     & 1.6      & 1.6   \\
\midrule
\multirow{3}{*}{Medium}         &   Halfcheetah    & \B 76.5(1.7)  & 49.1 & \B 77.6 & 61.6(15)   & 67.8(23)   & 42.1      & 47.2(06)     & 48.3     & 36.1  \\
                                &   Hopper         & \B 103.6(3.7) & 34.0 & 92.8    & 63.3(24)   & 20.9(139)  & 95.4      & 62.2(44)     & 59.3     & 29.0  \\
                                &   Walker2d       & \B 87.6(3.7)  & 72.1 & \B 86.9 & 70.1(50)   & -0.1(01)   & 77.8      & 76.1(16)     & 83.7     & 6.6   \\
\midrule                                
\multirow{3}{*}{Medium-Replay}  &   Halfcheetah    & \B 71.5(18)   & 57.5 & 68.9    & 57.0(12)   & 66.2(30)   & 40.2      & 44.6(07)     & 44.6     & 38.4  \\
                                &   Hopper         & \B 101.7(08)  & 62.2 & 96.6    & 71.5(80)   & 64.2(289)  & 93.6      & 98.3(13)     & 60.9     & 11.8  \\
                                &   Walker2d       & \B 87.7(30)   & 49.8 & 85.0    & 52.6(45)   & 67.9(157)  & 49.8      & 82.1(26)     & 81.8     & 11.3  \\
\midrule  
\multirow{3}{*}{Medium-Expert}  &   Halfcheetah    & \B 100.0(49)  & 55.4 & 93.7    & 65.3(97)       & 91.7(99)   & 53.3      & 90.6(44)     & 90.7     & 35.8  \\
                                &   Hopper         & \B 111.4(05)  & 82.6 & 83.3    & 105.4(45)  & 21.9(209)  & 108.7     & 98.2(110)    & 98.0     & \B 111.9 \\
                                &   Walker2d       & \B 112.3(18)  & 88.8 & 68.3    & 73.7(127)  & 4.0(54)    & 95.6      & 109.3(06)    & \B 110.1 & 6.4   \\
\midrule
\multicolumn{2}{c|}{MuJoCo-v2 Average:} & \B 78.8(20) & 51.2 & 68.9 & 55.9(44) & 37.6(92) & 64.4 & 62.0(25) & 58.2 & 25.1 \\
\bottomrule
\end{tabular}
\end{adjustbox}
\end{small}
\end{center}
\label{table:d4rl}
\end{table*}

\begin{table*}[t]
\caption{Performance of \texttt{Count-MORL} using each count estimation method~(\texttt{LC}, \texttt{AVG}, \texttt{UC}). We bold the highest score.}
\begin{center}
\begin{small}
\sisetup{separate-uncertainty}
\begin{adjustbox}{max width=\textwidth}
\begin{tabular}{
c
l|
S[detect-weight, table-format = 3.1(2)]
S[detect-weight, table-format = 3.1(2)]
S[detect-weight, table-format = 3.1(2)]
}
\toprule
 & & \multicolumn{3}{c}{Count Estimation} \\
{Dataset type} & {Environment} & {\texttt{LC} count} & {\texttt{AVG} count} & {\texttt{UC} count} \\
\midrule
\multirow{3}{*}{Random}         &   Halfcheetah    & \B 41.0(09) & 38.7(05) & 39.1(10)   \\
                                &   Hopper         & \B 30.7(13) & 27.7(62) & 26.5(65)   \\
                                &   Walker2d       & \B 21.9(01) & \B 21.9(01) & \B 21.9(02)   \\
\midrule  
\multirow{3}{*}{Medium}         &   Halfcheetah    & 74.2(25) & \B 76.5(17) & 74.6(16) \\
                                &   Hopper         & 99.7(72) & 101.8(47) & \B 103.6(37) \\
                                &   Walker2d       & 84.2(29) & \B 87.6(37) & 85.2(26)  \\
\midrule  
\multirow{3}{*}{Medium-Replay}  &   Halfcheetah    & 71.2(29) & 71.1(08) & \B 71.5(18) \\
                                &   Hopper         & 98.9(39) & 100.2(09) & \B 101.7(08) \\
                                &   Walker2d       & 84.3(31) & 85.8(28) & \B 87.7(30) \\
\midrule  
\multirow{3}{*}{Medium-Expert}  &   Halfcheetah    & 94.8(55)  & 98.1(25)  & \B 100.0(49) \\
                                &   Hopper         & 107.2(47) & 109.4(12) & \B 111.4(05) \\
                                &   Walker2d       & 109.7(14) & 110.7(05) & \B 112.3(18) \\
\bottomrule
\end{tabular}
\end{adjustbox}
\end{small}
\end{center}
\label{table:countestimation}
\end{table*}

Our experimental results, summarized in Tables~\ref{table:d4rl} and \ref{table:countestimation}, are based on evaluations carried out on the D4RL benchmark datasets. To address the question (ii), our method, \texttt{Count-MORL}, achieves the best or competitive performance in 10 out of the 12 settings. 
As seen in Table~\ref{table:d4rl}, \texttt{Count-MORL} outperforms others across datasets with narrower (\textit{Medium}, \textit{Medium-Expert}) and more diverse state-action distributions (\textit{Random}, \textit{Medium-Replay}). 
However, the \textit{Random} datasets of \textit{Hopper} and \textit{Walker2d} are exceptions, where MOReL performs better. Excluding MOReL from comparison, our method leads in performance for \textit{Hopper} and \textit{Walker2d} on the \textit{Random} dataset. Additionally, our method surpasses all other algorithms for the \textit{Random} dataset of \textit{Halfcheetah}. These results demonstrate the superior performance of \texttt{Count-MORL} against state-of-the-art offline RL algorithms across various dataset types.

Table~\ref{table:countestimation} exhibits the performance of different count estimation methods (\texttt{LC}, \texttt{AVG}, and \texttt{UC}) on the D4RL benchmark datasets. 
\texttt{LC} count performs best for \textit{Random} datasets, while \texttt{UC} count outperforms others for \textit{Medium-Expert} datasets. These results suggest that we can tailor the count estimation method based on the nature of the offline dataset. Unobserved state-action pairs in the offline dataset are assigned a constant penalty of $1$ (see Corollary~\ref{cor:errorbound}). 
For the \textit{Random} dataset, \texttt{LC} count provides a relatively larger penalty to observed state-action pairs by under-estimating counts, 
allowing the agent to potentially take actions or visit states beyond the dataset's boundaries.
Conversely, for the \textit{Medium-Expert} dataset, \texttt{UC} count computes over-approximate counts
incurring smaller penalties and encouraging the agent to exploit the states within the dataset. 
If the replay buffer consistently contains high-quality data, as observed in the \textit{Medium-Replay} datasets, \texttt{UC} count tends to perform better, as indicated in Table~\ref{table:countestimation}. 
However, if the behavior policy generates low-quality data significantly different from data from an optimal policy, \texttt{LC} count can be more beneficial. 
Thus, 
the appropriate choice of count estimation is likely to achieve improved performance.

\section{Conclusion}
\texttt{Count-MORL} introduces a novel and tractable approach to model-based offline RL that incorporates count-based conservatism, effectively bridging the theoretical and practical divide in model-based offline deep RL. 
Consequently, our method outperforms existing state-of-the-art offline RL algorithms, further solidifying our approach's efficacy.

\section*{Acknowledgements}
This work was supported by the National Research Foundation of Korea(NRF) grant funded by the Korea government(MSIT) (No. 2021M3E5D2A01024795, No. 2022R1C1C1006859, and No.~2022R1A4A103057912).

\bibliography{reference}
\bibliographystyle{icml2023}

\newpage
\appendix
\onecolumn
\section{Proosfs}
\subsection{Proof of \cref{thm:estimationerror}}
\begin{proof}
    Given a state-action $(s,a)$ pair observed in $\gD$ with $\gD_{s, a} =\{(s_{i}, a_{i}, s'_{i})\}_{s_{i} = s, a_{i} = a}$ and $\cnt = |\gD_{s, a}|$ such that the MLE of transition dynamics is
    \begin{equation}
        \label{eq:MLEdynamics}
        \lP(\cdot \mid s, a) \in \argmax_{P \in \gM} \sum_{(s,a,s') \in \gD_{s, a}} \log P(s' \mid s, a)
    \end{equation}
    for given $(s,a)$.
    By Theorem 21 in \citet{agarwal2020flambe}, with probability at least $1-\delta$,
    \begin{align*}
        \mathbb{E}_{(s_{i},a_{i}) \sim \gD_{s,a}} \left[ \mathrm{TV}\!\left(\lP(\cdot \mid s_{i},a_{i}), P^{\star}(\cdot \mid s_{i},a_{i}) \right)^{2} \right]
        \le
       \frac{2\log(|\gM|/\delta)}{n\pare} \, .
    \end{align*}
    We can directly bound the total variation distance between the estimated transition dynamics $\lP$ and the true transition dynamics $P^{\star}$ due to the subset $\gD_{s, a} =\{(s, a, s'_{i})\}_{i=1}^{n(s, a)}$.
    Thus,
    \begin{equation*}
         \mathrm{TV}\!\left(\lP(\cdot \mid s,a), P^{\star}(\cdot \mid s,a) \right) \le \sqrt{\frac{2\log(|\gM|/\delta)}{n\pare}}.
    \end{equation*}
\end{proof}

\subsubsection{Proof of Corollary~\ref{cor:errorbound}}
\begin{proof}
    By \cref{thm:estimationerror}, given a state-action $(s,a)$ pair observed in $\gD$, the estimated transition dynamics is satisfied by \cref{eq:MLEdynamics} such that we have
    \begin{equation*}
         \mathrm{TV}\!\left(\lP(\cdot \mid s,a), P^{\star}(\cdot \mid s,a) \right) \le \sqrt{\frac{2\log(|\gM|/\delta)}{n\pare}}.
    \end{equation*}
    If given an unobserved state-action pair in $\gD$, we cannot estimate the transition dynamics or compute the estimation error.
    And, when $n(s, a)$ is less than $\sqrt{2 \log (|\gM|/\delta)}$, the estimation error becomes greater than $1$.
    Thus, by the definition of the total variation distance between probability distributions $P$ and $Q$ as
    \begin{equation*}
        \mathrm{TV}\!\left(P, Q \right) = \sup_{A} \lvert P(A) - Q(A) \rvert,
    \end{equation*}
    we bound the estimation error as $1$ for these cases.
    Therefore, with probability at least $1 - \delta$, for any state-action pair $\pare \in \gS \times \gA$,
    \begin{equation*}
        \mathrm{TV}\!\left(\lP(\cdot \mid s,a), P^{\star}(\cdot \mid s,a)\right)
        \le
        \min \left(1, \sqrt{\frac{2\log(|\gM|/\delta)}{\cnt}} \right).
    \end{equation*}
\end{proof}

\subsection{Proof of \cref{thm:sub-optimalitygap}}
Before we prove \cref{thm:sub-optimalitygap}, we prove \cref{lemma:valuegap} and \cref{cor:conservatismMDP} that will be used in the latter proof.

\subsubsection{Proof of \cref{lemma:valuegap}}
The proof of \cref{lemma:valuegap} follows from modifying the proofs of Lemma~4.1 in MOPO~\citep{yu2020mopo}
based on count-based estimation error.
\begin{proof}
    We denote the expectation of cumulative discounted reward under $\pi$ and $\lP$ for $j$ steps and then switching to $P^{\star}$ for the rest steps:
    \begin{equation*}
        W_{j} = 
        \underset{\substack{s \sim d_{0} \\
        \forall t \ge 0, a_{t} \sim \pi(\cdot \mid s_{t}) \\
        \forall j > t \ge 0, s_{t+1} \sim \lP(\cdot \mid s_{t}, a_{t}) \\ 
        \forall t \ge j, s_{t+1} \sim P^{\star}(\cdot \mid s_{t}, a_{t})}}
        {\mathbb{E}}
        \left[ \sum_{t=0}^{\infty} \gamma^{t} r(s_{t}, a_{t}) \mid s_{0} = s \right]
    \end{equation*}
    Note that $W_{0} = V_{M^{\star}}^{\pi}$ and $W_{\infty} = V_{\lM}^{\pi}$.
    By the telescoping lemma, we have that
    \begin{equation*}
        V_{\lM}^{\pi} - V_{M^{\star}}^{\pi}
        =
        \sum_{j=0}^{\infty} \left( W_{j+1} - W_{j} \right).
    \end{equation*}
    Rewriting $W_{j}$ and $W_{j+1}$ for the trajectory distribution as
    \begin{align*}
        W_{j} &= R_{j} +\underset{s_{j},a_{j} \sim \pi,\lP}{\mathbb{E}} \left[ \underset{s_{j+1} \sim \lP(\cdot \mid s_{j}, a_{j})}{\mathbb{E}} \left[ \gamma^{j+1} V_{P^{\star},r}^{\pi}(s_{j+1}) \right] \right] \\
        W_{j+1} &= R_{j} +\underset{s_{j},a_{j} \sim \pi,\lP}{\mathbb{E}} \left[ \underset{s_{j+1} \sim P^{\star}(\cdot \mid s_{j}, a_{j})}{\mathbb{E}} \left[ \gamma^{j+1} V_{P^{\star},r}^{\pi}(s_{j+1}) \right] \right],
    \end{align*}
    where $R_{j}$ denotes the sum of the discounted rewards obtained from the first step to $j$ step under $\pi$ and $\lP$.
    
    Then,
    \begin{align}
        W_{j+1} - W_{j} \nonumber
        &=
        \gamma^{j+1}
        \underset{s_{j},a_{j} \sim \pi,\lP}{\mathbb{E}} \left[ \underset{s_{j+1} \sim \lP(\cdot \mid s_{j}, a_{j})}{\mathbb{E}} \left[  V_{P^{\star},r}^{\pi}(s_{j+1}) \right] - \underset{s_{j+1} \sim P^{\star}(\cdot \mid s_{j}, a_{j})}{\mathbb{E}} \left[ V_{P^{\star},r}^{\pi}(s_{j+1}) \right] \right] \nonumber\\
        &\le
        \frac{\gamma^{j+1} R_{\mathrm{max}}}{1-\gamma}
        \underset{s_{j},a_{j} \sim \pi,\lP}{\mathbb{E}} \left[ \mathrm{TV}\!\left(\lP(\cdot|s_{j},a_{j}), P^{\star}(\cdot|s_{j},a_{j})\right) \right] \nonumber \\
        &\le
        \frac{\gamma^{j+1} R_{\mathrm{max}}}{1-\gamma}
        \underset{s_{j},a_{j} \sim \pi,\lP}{\mathbb{E}} \left[ C_{\lP}^{\delta}(s_{j},a_{j}) \right] \nonumber \\
        &\le
        \frac{\gamma^{j+1} R_{\mathrm{max}}}{1-\gamma}
        \underset{s_{j},a_{j} \sim \pi,\lP}{\mathbb{E}} \left[ \widehat{C}_{\lP}^{\delta}(s_{j},a_{j}) + \epsilon(s_{j}, a_{j}) \right] \label{eq:ww},
    \end{align}
    where first inequality step uses the fact that $\left\lVert V_{P^{\star}, r}^{\pi}(s) \right\rVert_{\infty} \le \frac{R_{\mathrm{max}}}{1-\gamma}$;
    second inequality step uses the estimation error bound based on the true count $C_{\lP}^{\delta}$ in \cref{cor:errorbound};
    and the last inequality step uses the estimated error bound based on the approximate count $\widehat{C}_{\lP}^{\delta}$
    and denote the approximation error for each state-action pair as $\epsilon(s,a) := \left| C_{\lP}^{\delta}\pare - \widehat{C}_{\lP}^{\delta}\pare \right|$.

    Therefore, with probability at least $1 - \delta$, for any policy $\pi$,
    \begin{align*}
        V_{\lM}^{\pi} - V_{M^{\star}}^{\pi}
        &=
        \sum_{j=0}^{\infty} \left( W_{j+1} - W_{j} \right) \\
        &\le
        \frac{R_{\mathrm{max}}}{1-\gamma} \sum_{j=0}^{\infty} \gamma^{j+1}
        \underset{s_{j},a_{j} \sim \pi,\lP}{\mathbb{E}} \left[ \widehat{C}_{\lP}^{\delta}(s_{j},a_{j}) + \epsilon(s_{j}, a_{j}) \right] \\
        &=
        \frac{\gamma R_{\mathrm{max}}}{(1-\gamma)^{2}}
        \mathbb{E}_{\pare \sim d_{\lP}^{\pi}} \left[ \widehat{C}_{\lP}^{\delta}\pare + \epsilon(s_{j}, a_{j}) \right] \\
        &\le
        \frac{\gamma R_{\mathrm{max}}}{(1-\gamma)^{2}} \mathbb{E}_{\pare \sim d_{\lP}^{\pi}} \left[ \widehat{C}_{\lP}^{\delta}\pare \right]
        + \frac{\gamma R_{\mathrm{max}}}{(1-\gamma)^{2}} \epsilon~,
    \end{align*}
    where the first inequality step uses \cref{eq:ww}; and the last inequality uses the definition of the maximal approximation error as $\epsilon = \sup_{(s, a) \in \gS \times \gA} \epsilon(s, a)$.
\end{proof}

\subsubsection{Proof of \cref{cor:conservatismMDP}}
\begin{proof}
    In \cref{def:conservative-MDP}, we define the count-based conservatism MDP $\pM := (\gS, \gA, \lP, \pR, d_{0}, \gamma)$ with the estimated transition dynamics $\lP$ and the count-based penalized reward $\pR\pare = r\pare - \frac{\gamma R_{\mathrm{max}}}{1-\gamma} \widehat{C}_{\lP}^{\delta}\pare$.
    Then, we represent the result of \cref{lemma:valuegap} based on the value gap between the count-based conservatism MDP $\pM$ and the true MDP $M^{\star}$.
    \begin{align}
        V_{M^{\star}}^{\pi} \nonumber
        &\ge
        V_{\lM}^{\pi}
        - \frac{\gamma R_{\mathrm{max}}}{(1-\gamma)^{2}} \mathbb{E}_{\pare \sim d_{\lP}^{\pi}} \left[ \widehat{C}_{\lP}^{\delta}\pare \right]
        - \frac{\gamma R_{\mathrm{max}}}{(1-\gamma)^{2}} \epsilon \nonumber \\
        &=
        \frac{1}{1-\gamma} \mathbb{E}_{\pare \sim d_{\lP}^{\pi}} \left[ r\pare - \frac{\gamma R_{\mathrm{max}}}{1-\gamma}\widehat{C}_{\lP}^{\delta}\pare \right]
        - \frac{\gamma R_{\mathrm{max}}}{(1-\gamma)^{2}} \epsilon \nonumber \\
        &=
        \frac{1}{1-\gamma} \mathbb{E}_{\pare \sim d_{\lP}^{\pi}} \left[ \pR\pare \right]
        - \frac{\gamma R_{\mathrm{max}}}{(1-\gamma)^{2}} \epsilon \nonumber \\
        &= V_{\pM}^{\pi} - \frac{\gamma R_{\mathrm{max}}}{(1-\gamma)^{2}}\epsilon \label{eq:conservativegap}~,
    \end{align}
    where the first inequality step uses the value gap of the estimated model in \cref{lemma:valuegap}; and the second equality step uses the definition of the count-based penalized reward. 
\end{proof}

Now, we prove our main theorem (\cref{thm:sub-optimalitygap}) that provides a performance guarantee under $M^{\star}$.
\begin{proof}
    For any policy $\pi$, we have that
    \begin{align*}
         V_{M^{\star}}^{\hat{\pi}}
         &\ge
         V_{\pM}^{\hat{\pi}} - \frac{\gamma R_{\mathrm{max}}}{(1-\gamma)^{2}}\epsilon \\
         &\ge
         V_{\pM}^{\pi} - \frac{\gamma R_{\mathrm{max}}}{(1-\gamma)^{2}}\epsilon \\
         &=
         V_{\lM}^{\pi} - \frac{\gamma R_{\mathrm{max}}}{(1-\gamma)^{2}} \mathbb{E}_{\pare \sim d_{\lP}^{\pi}} \left[ \widehat{C}_{\lP}^{\delta}\pare \right] - \frac{\gamma R_{\mathrm{max}}}{(1-\gamma)^{2}}\epsilon \\
         &\ge
         V_{M^{\star}}^{\pi} - \frac{2\gamma R_{\mathrm{max}}}{(1-\gamma)^{2}} \mathbb{E}_{\pare \sim d_{\lP}^{\pi}} \left[ \widehat{C}_{\lP}^{\delta}\pare \right] - \frac{2\gamma R_{\mathrm{max}}}{(1-\gamma)^{2}}\epsilon ~,
    \end{align*}
    where the first inequality step uses the result of \cref{cor:conservatismMDP}; the second inequality step uses the definition of $\hat{\pi}$ as $\hat{\pi} = \argmax_{\pi}V_{\pM}^{\pi}$; and the last inequality step uses the result of \cref{thm:estimationerror}.
\end{proof}

\section{Experimental Details.}
This section contains all details of \texttt{Count-MORL} (\href{https://github.com/oh-lab/Count-MORL}{github.com/oh-lab/Count-MORL}) with hash codes based on the official MOPO code from \href{https://github.com/tianheyu927/mopo}{github.com/tianheyu927/mopo}.

\subsection{Model and Policy Training}
We represent the dynamics model as an ensemble of probabilistic neural networks that outputs a Gaussian distribution over the next state and reward given the current state and action:
\begin{equation*}
    \lP_{\theta}(s',r|s,a) = \gN (\mu_{\theta}(s,a), \Sigma_{\theta}(s,a)).
\end{equation*}
Each dynamics model consists of a $4$-layer neural network with 200 hidden units per layer and after the last hidden layer, the dynamics model outputs the mean and variance using a two-head architecture.
We connect an autoencoder to the hidden layer of the dynamics model, which takes the output of this hidden layer as an input.
We determine the architecture of the autoencoder with the output dimension $d$ of the bottleneck layer.
For $d<50$, we use 6-layer neural network with $[100, 50, d, 50, 100, 100]$ units.
For $50 \le d < 100$, we use 5-layer neural network with $[100, 100, d, 100, 100]$ units.
We learn the dynamics model and the autoencoder using the log-likelihood objective function and mean square error, respectively.
Following previous works \citep{janner2019trust, yu2020mopo, yu2021combo, rigterrambo}, we train an ensemble of $7$ such models that each contain the dynamics model and autoencoder and pick the best $5$ models based on the validation prediction error on a held-out test set of $1000$ transitions from the offline dataset $\gD$.
For the soft actor-critic (SAC)~\citep{haarnoja2018soft} updates, we sample a batch of 256 transitions, $5$\% of them from $\gD$ and the rest of them from $\gD_{\mathrm{model}}$.

\subsection{Hyperparameters}
We found that the hyperparameters that have a significant influence on the performance of \texttt{Count-MORL}.
We take a rollout length ($H$), a standard deviation coefficient ($\alpha$) for the count estimation, a reward penalty coefficient ($\beta$) for each count estimation method, and a dimension of hash codes ($d$).
For a rollout length and a reward penalty coefficient, we found that a length $H \in \{5, 20\}$ and a coefficient $\beta \in \{0.5, 1, 3, 5\}$ performed well across all datasets.
This is a slight modification to the values of $H \in \{1, 5\}$ and $\beta \in \{0.5, 1, 5\}$ in previous model-based offline RL algorithms~\citep{yu2020mopo}.
And, in \citet{lurevisiting}, the authors show that a rollout length and a reward penalty coefficient play key parameters in determining the performance of model-based offline RL algorithms.
Thus, we utilize the longer rollout length as $20$ steps.
We fix a standard deviation coefficient $\alpha$ as $0.5$.

Depending on the dataset, we take a dimension $d$ of hash code over $\{16, 32, 50, 64, 80\}$.
We experimentally confirmed that the number of hash codes used for counting samples in the offline dataset does not exponentially increase when the binary vector has a high dimension.
For each dimension of the hash code, the count of samples ranged from 5 to 10 (the total number of samples in the offline dataset divided by the number of hash codes used for counting) in the highest reported score for each data type on average.
In some other cases, the sample count for each hash code ranged from 1 to 2, or even up to 100.
Details of implementation are in the \cref{table:hyperparameters}.

\begin{table*}[ht]
\caption{Hyperparameters used in the D4RL datasets.}
\begin{center}
\begin{small}
\begin{adjustbox}{max width=\textwidth}
\begin{tabular}{cl|c|ccc|c}
\toprule
 &  & &  \multicolumn{3}{c|}{$\beta$} &   \\
{Dataset type} & {Environment} & $H$ & {\texttt{LC}} & {\texttt{AVG}} & {\texttt{UC}} & $d$ \\
\midrule
\multirow{3}{*}{Random}         &   Halfcheetah    & 5  & 1  & 0.5 & 1 & 50   \\
                                &   Hopper         & 20 & 1  & 1   & 3 & 80  \\
                                &   Walker2d       & 20 & 1  & 1   & 1 & 64   \\
\midrule 
\multirow{3}{*}{Medium}         &   Halfcheetah    & 5  & 1  & 1   & 3 & 32 \\
                                &   Hopper         & 20 & 1  & 1   & 1 & 50  \\
                                &   Walker2d       & 20 & 3  & 3   & 3 & 32  \\
\midrule 
\multirow{3}{*}{Medium-Replay}  &   Halfcheetah    & 5  & 1  & 3   & 3 & 32 \\
                                &   Hopper         & 5  & 3  & 3   & 1 & 50 \\
                                &   Walker2d       & 5  & 1  & 3   & 1 & 32 \\
\midrule
\multirow{3}{*}{Medium-Expert}  &   Halfcheetah    & 5  & 3  & 1   & 3 & 32 \\
                                &   Hopper         & 20 & 3  & 3   & 3 & 50 \\
                                &   Walker2d       & 20 & 3  & 3   & 3 & 50 \\
\bottomrule
\end{tabular}
\end{adjustbox}
\end{small}
\end{center}
\label{table:hyperparameters}
\end{table*}

\section{Additional Results}
\subsection{Results on Grid-World}
\label{appendix:grid-world}
We compute the approximate count and the known true count on the 8$\times$8 Grid-World environments.
Each environment comprises $256$ state-action pairs from $64$ states, each with $4$ actions~(up, down, left, right).
We convert all state-action pairs into the multi-hot encoder to input. 
Before counting the number of samples, we train the dynamics model and the autoencoder with the $20$-dimension of the bottleneck layer on the replay buffer of the policy trained by Q-learning.
When we use the bottleneck layer's dimension less than $20$, the hash code cannot divide samples into $256$ clusters since the rounding of the bottleneck layer's output turns to the value either $0$ or $1$.

\begin{figure}[t]
    \centering
    \begin{subfigure}[b]{0.475\textwidth}
        \includegraphics[width=8cm]{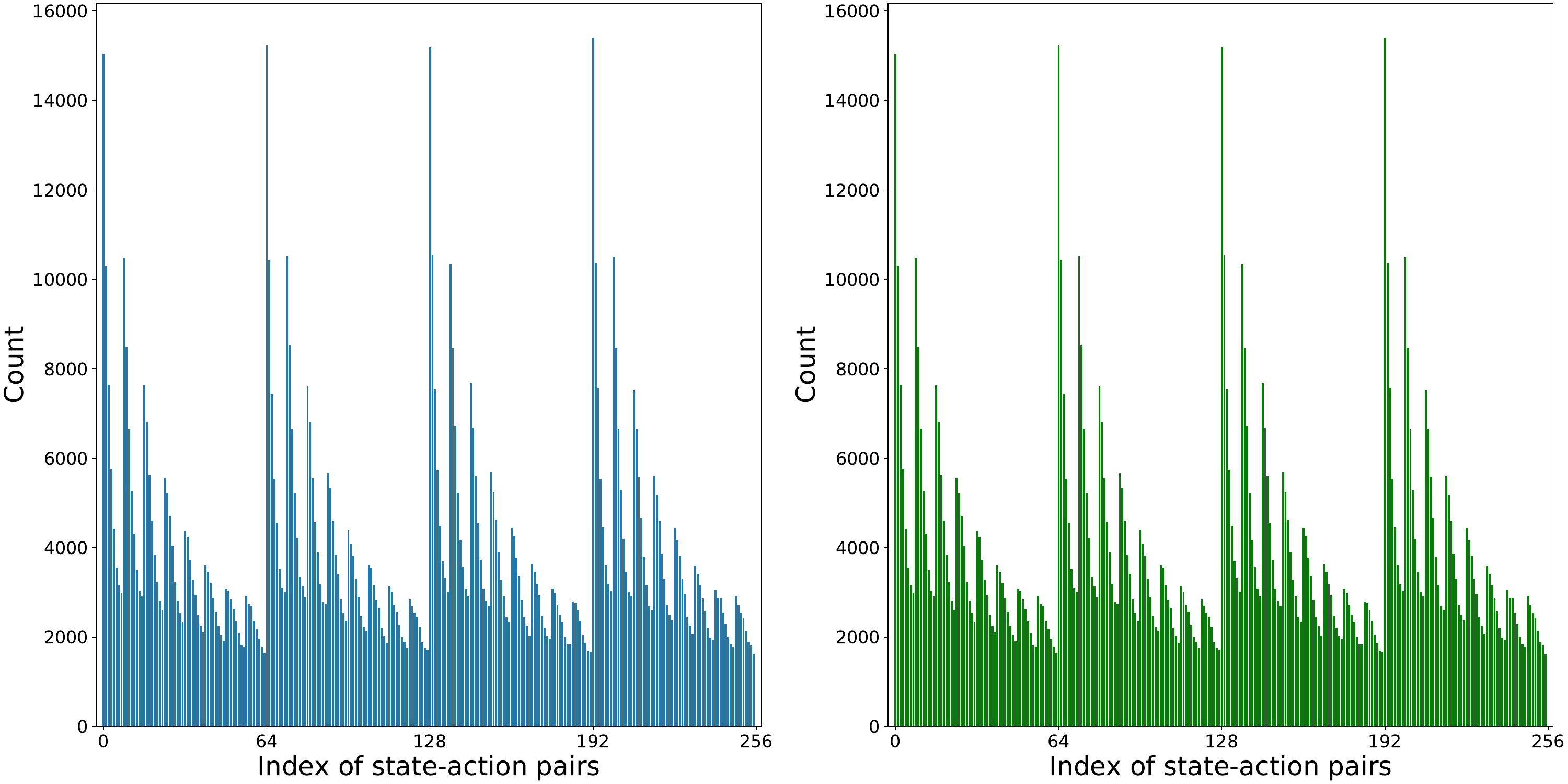}
        \caption{Empty}
    \end{subfigure}
    \hfill
    \vspace{0.5cm}
    \begin{subfigure}[b]{0.475\textwidth}
        \includegraphics[width=8cm]{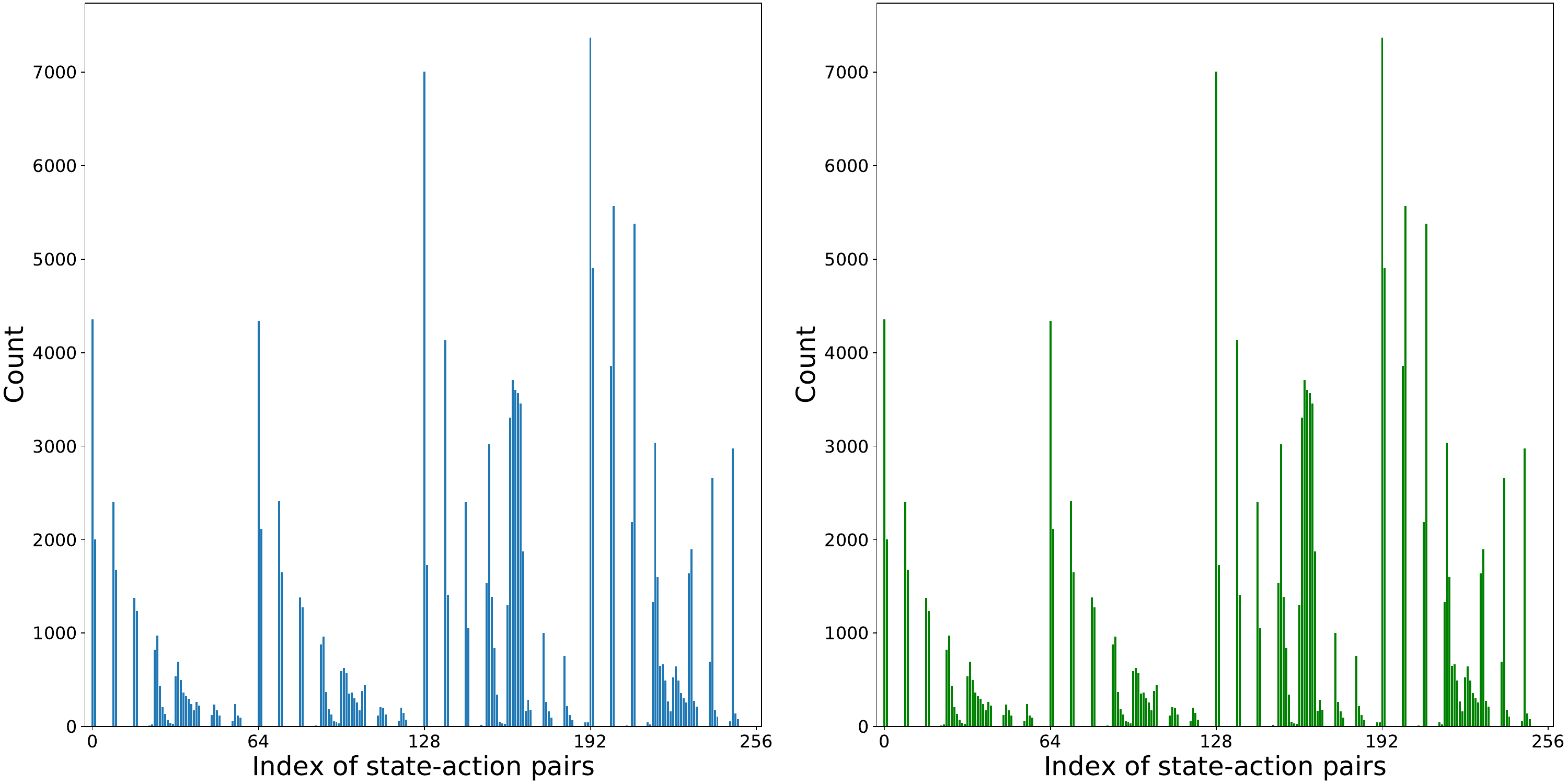}
        \caption{Bridge}
    \end{subfigure}
    \hfill
    \begin{subfigure}[b]{0.475\textwidth}
        \includegraphics[width=8cm]{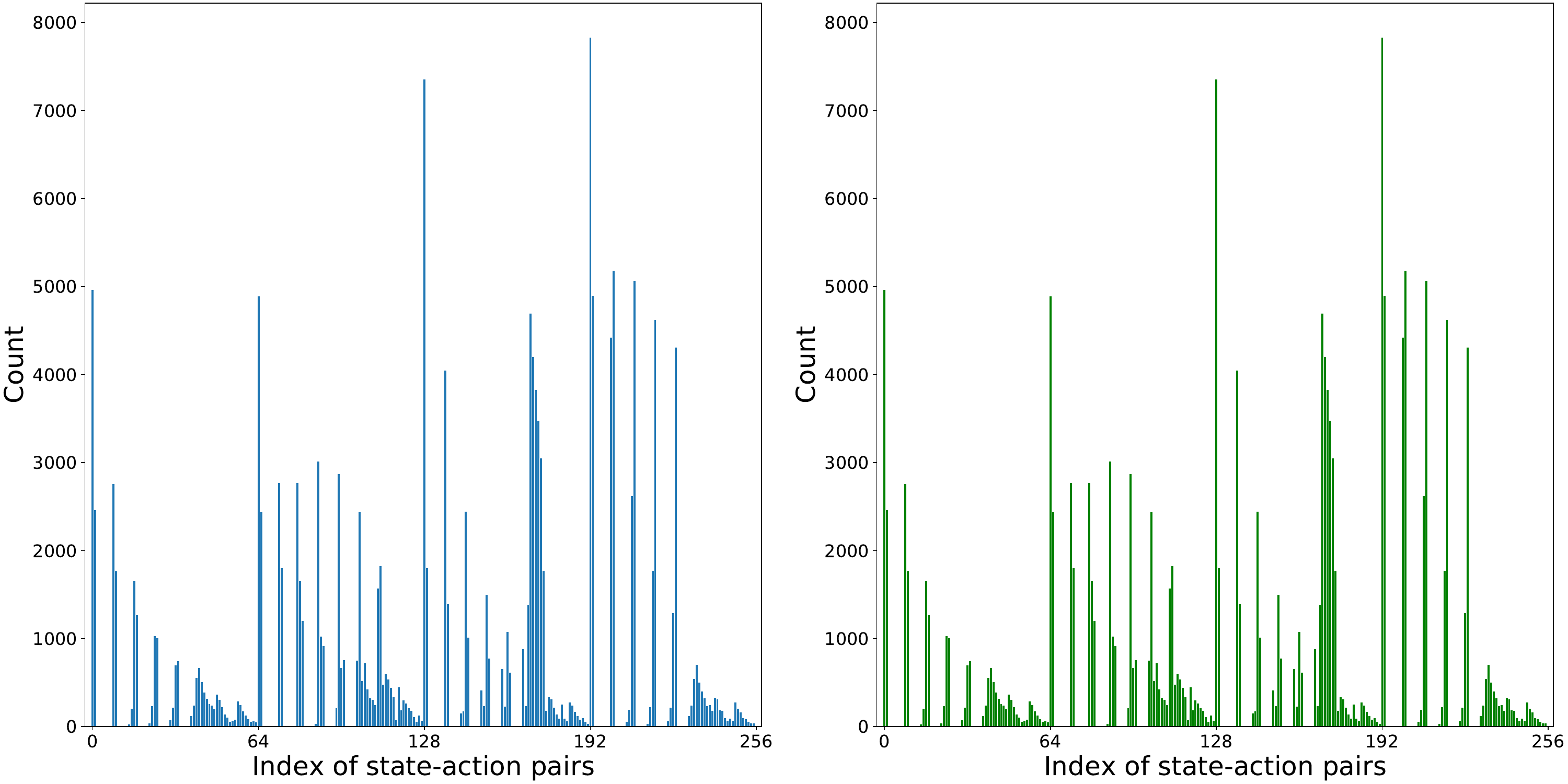}
        \caption{Cliff}
    \end{subfigure}
    \hfill
    \begin{subfigure}[b]{0.475\textwidth}
        \includegraphics[width=8cm]{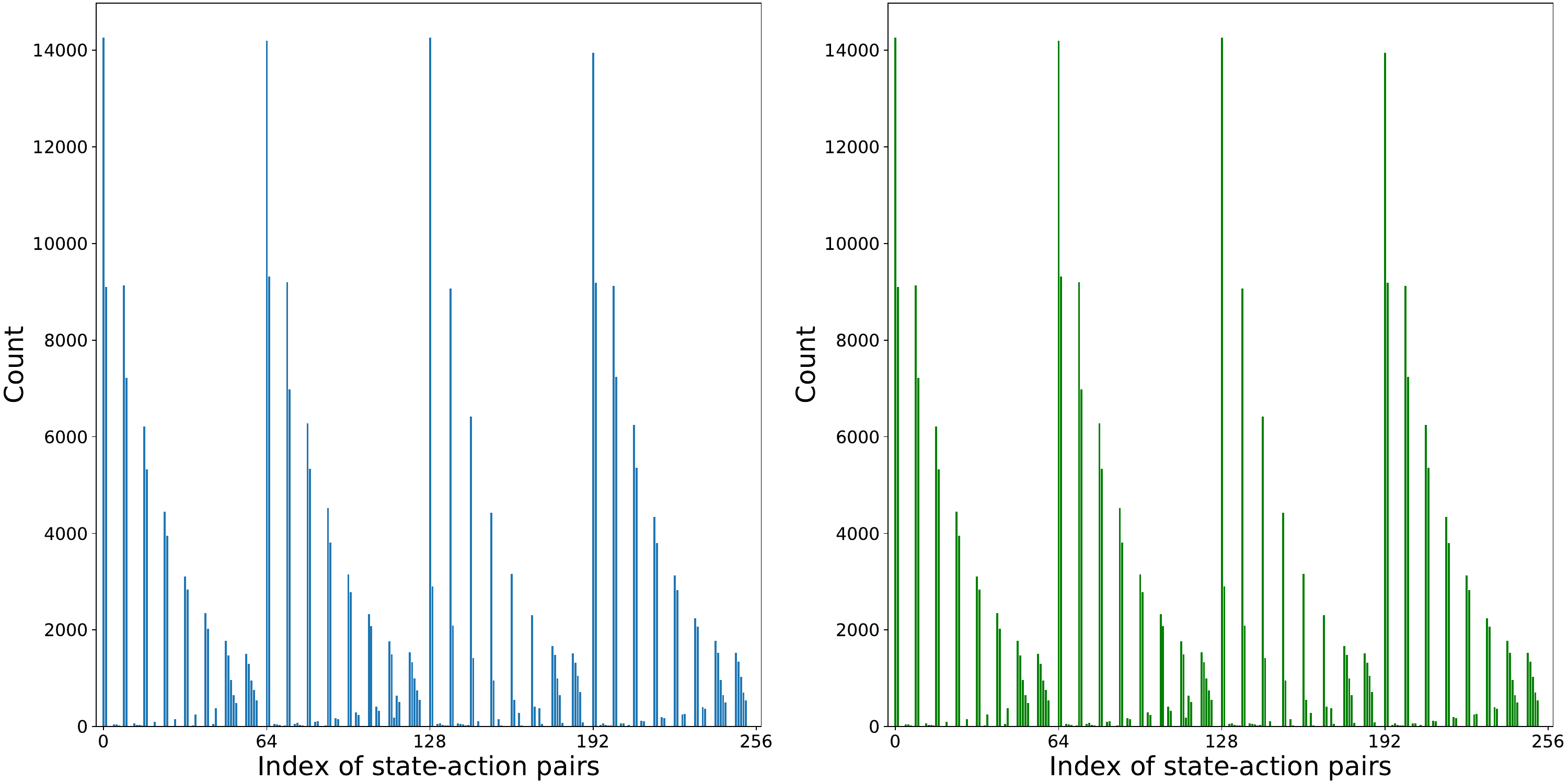}
        \caption{Zigzag}
    \end{subfigure}
\caption{Comparison between the true count~(\textcolor[rgb]{0, 0.44, 0.75}{Blue}) and the approximate count~(\textcolor[rgb]{0.44, 0.678, 0.2745}{Green}) on Grid-World environments.}
\label{fig:estimatedcount}
\end{figure}

\textit{Empty}, \textit{Bridge}, \textit{Cliff}, and \textit{ZigZag} datasets contain $10^{6}$, $1.6\times 10^{5}$, $1.4 \times 10^{5}$ and $3 \times 10^{5}$ transition samples for each.
\textit{Empty} dataset has samples for all state-action pairs, but \textit{Bridge}, \textit{Cliff} and \textit{Zigzag} datasets have no samples on all actions taken in the lava~(black grids).
In \cref{fig:estimatedcount}, the blue and green histogram presents the known true count and the approximate count, respectively.
We observe that the count error between the true and approximate count is zero for all state-action pairs on the four environments.
Therefore, \cref{fig:estimatedcount} shows that our implementation model structure is able to exactly estimate the true count in Grid-World.

\subsection{Dimension of hash codes}
In \cref{table:dimension}, we investigate the impact of hash code dimension on performance.
We perform a rough grid search over a range of the dimension of hash codes.
In fact, without much intensive the dimension of hash codes tuning to derive the reported results, \texttt{Count-MORL} is shown to perform significantly superior to the existing offline deep RL algorithms, which we believe was another strength of our method.
We implement experiments on a total of five dimensions for each dataset, evaluating count estimation methods that showed good performance in \cref{table:countestimation}.

\begin{table*}[h]
\caption{Performance of \texttt{Count-MORL} for each dimension of hash codes. We bold the highest score.}
\begin{center}
\begin{small}
\sisetup{separate-uncertainty}
\begin{adjustbox}{max width=\textwidth}
\begin{tabular}{
c
l|
c|
S[detect-weight, table-format = 3.1(2)]
S[detect-weight, table-format = 3.1(2)]
S[detect-weight, table-format = 3.1(2)]
S[detect-weight, table-format = 3.1(2)]
S[detect-weight, table-format = 3.1(2)]
}
\toprule
 & &  & \multicolumn{5}{c}{Dimension of hash codes} \\
{Dataset type} & {Environment} & {Count} & {$d=16$} & {$d=32$} & {$d=50$} & {$d=64$} & {$d=80$} \\
\midrule
\multirow{3}{*}{Random}         &   Halfcheetah    & \texttt{LC} & 36.7(05) & 38.4(09) & \B 41.0(09) & 36.2(10) & 35.6(08) \\
                                &   Hopper         & \texttt{LC} & 22.9(62) & 21.5(84) & 24.7(68) & 27.6(64) & \B 30.7(13) \\
                                &   Walker2d       & \texttt{LC} & 21.8(01) & 21.8(01) & \B 21.9(01) & \B 21.9(01) & 21.8(01) \\
\midrule  
\multirow{3}{*}{Medium}         &   Halfcheetah    & \texttt{AVG} & 74.4(15) & \B 76.5(17) & 75.7(14) & 75.2(19) & 73.9(16) \\
                                &   Hopper         & \texttt{UC} & 90.5(31)  & 93.2(46) & \B 103.6(37) & 98.3(58) & 95.1(39) \\
                                &   Walker2d       & \texttt{AVG} & 82.2(09) & \B 87.6(37) & 81.4(18) & 80.1(15) & 80.5(07)  \\
\midrule  
\multirow{3}{*}{Medium-Replay}  &   Halfcheetah    & \texttt{UC} & 68.7(19)  & \B 71.5(18) & 68.2(15) & 67.3(17) & 65.8(17) \\
                                &   Hopper         & \texttt{UC} & 94.4(37)  & 97.1(41) & \B 101.7(08) & 94.6(49) & 92.3(32) \\
                                &   Walker2d       & \texttt{UC} & 85.8(17)  & \B 87.7(30) & 82.4(38) & 80.7(13) & 81.0(25) \\
\midrule  
\multirow{3}{*}{Medium-Expert}  &   Halfcheetah    & \texttt{UC} & 98.1(34)  & \B 100.0(49) & 99.1(28) & 98.6(32) & 93.4(27) \\
                                &   Hopper         & \texttt{UC} & 95.2(73)  & 90.3(16.6)  & \B 111.4(05) & 108.3(18) & 102.8(32) \\
                                &   Walker2d       & \texttt{UC} & 103.4(26) & 106.1(34) & \B 112.3(18) & 109.6(20) & 105.8(29) \\
\bottomrule
\end{tabular}
\end{adjustbox}
\end{small}
\end{center}
\label{table:dimension}
\end{table*}

\subsection{Performance on MuJoCo-v2 datasets}
\begin{figure}[h]
    \centering
    \includegraphics[width=0.97\textwidth,trim={7cm 9cm 7cm 4.5cm}, clip]{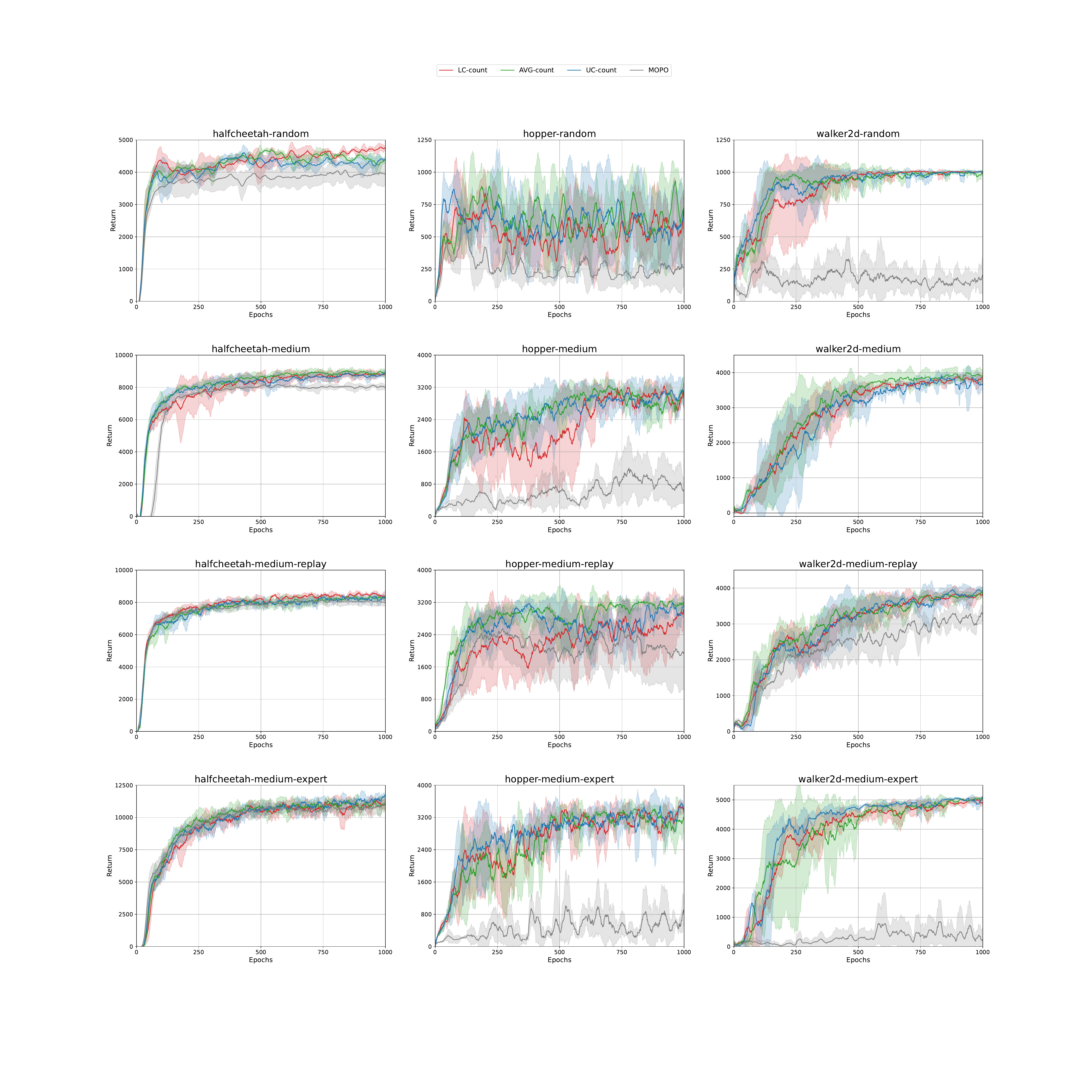}
    \caption{Performance of MuJoCo-v2 datasets}
    \label{fig:performance-v2}
\end{figure}

We shows the performance of \texttt{Count-MORL}~(\texttt{LC}, \texttt{AVG}, \texttt{UC}) and MOPO on MuJoCo-v2 datasets.
We confirm that our algorithm performs better than MOPO just by applying the count-based conservatism instead of the uncertainty heuristics of the model in \cref{fig:performance-v2}.
The $x$-axis of the graph represents episodes, while the $y$-axis represents the cumulative reward.
All count estimation methods are able to train the estimated policy with fewer episodes compared to MOPO to approximate an optimal policy.

\subsection{Results on MuJoCo-v0 datasets}
In Table~\ref{table:d4rl}, we show the performance of \texttt{Count-MORL} and offline RL baselines on MuJoCo-v2 datasets.
When reproducing the performance in MuJoCo-v2 datasets, we used the implementation of COMBO from \href{https://github.com/takuseno/d3rlpy}{github.com/takuseno/d3rlpy}, MOPO from \href{https://github.com/tianheyu927/mopo}{github.com/tianheyu927/mopo}, and CQL from \href{https://github.com/yihaosun1124/OfflineRL-Kit}{github.com/yihaosun1124/OfflineRL-Kit}.
However, the reproduced scores for MOPO, COMBO, and CQL are lower than those from their paper.
Therefore, we implement \texttt{Count-MORL} with hash codes on MuJoCo-v0 datasets in \cref{table:mujoco-v0}.
\texttt{Count-MORL} achieves the best or comparable performance on $10$ out of $12$ settings in MuJoCo-v0 datasets.

\begin{table*}[h]
\caption{Results for D4RL datasets. Each number is the normalized score proposed in \citealt{fu2020d4rl} of the policy during the last $5$ iterations of training averaged over $5$ seeds,
where $\pm$ denotes the standard deviation over seeds. We take the results of COMBO, MOPO and CQL from their original papers. We bold the scores within 2\% of the highest score across all algorithms.}
\begin{center}
\begin{small}
\sisetup{separate-uncertainty}
\begin{adjustbox}{max width=\textwidth}
\begin{tabular}{
c
l|
S[detect-weight, table-format = 3.1(2)]|
c
c
c
}
\toprule
{Dataset type} & {Environment} & {\myalg} & {COMBO} & {MOPO} & {CQL} \\
\midrule
\multirow{3}{*}{Random}         &   Halfcheetah    & \B 40.5(04)    & 38.8       &35.4       &35.4    \\
                                &   Hopper         & 11.9(02)       & \B 17.9       &11.7       &10.8     \\
                                &   Walker2d       & \B 21.5(02)    & 7.0        &13.6       &7.0     \\
\midrule
\multirow{3}{*}{Medium}         &   Halfcheetah    & \B 56.2(08)    & 54.2       &42.3       &44.4    \\
                                &   Hopper         &  82.3(2.4)     & \B 97.2       &28.0       &86.6    \\
                                &   Walker2d       & \B 80.5(0.7)   & \B 81.9       &17.8       &74.5     \\
\midrule                                
\multirow{3}{*}{Medium-Replay}  &   Halfcheetah    & \B 59.4(05)    & 55.1       &53.1       &46.2    \\
                                &   Hopper         & \B 91.1(07)    & \B 89.5       &67.5       &48.6    \\
                                &   Walker2d       & \B 76.1(40)    & 56.0       &39.0       &32.6    \\
\midrule  
\multirow{3}{*}{Medium-Expert}  &   Halfcheetah    & \B 102.9(01)   & 90.0       &63.3       &62.4     \\
                                &   Hopper         & \B 112.1(03)   & \B 111.1      &23.7       &\B 111.0   \\
                                &   Walker2d       & \B 102.3(05)   & \B 103.3      &44.6       &98.7    \\
\midrule
\multicolumn{2}{c|}{MuJoCo-v0 Average:} & 69.7(09) & 66.8 & 36.7  & 54.9 \\
\bottomrule
\end{tabular}
\end{adjustbox}
\end{small}
\end{center}
\label{table:mujoco-v0}
\end{table*}

\end{document}